\documentclass[journal,10pt]{IEEEtran}

\usepackage{tikz}
\usepackage{pgfplots}
\usepackage{tkz-euclide}
\usepackage{cite}
\usepackage{amsmath,amssymb,amsfonts}
\usepackage{algorithmic}
\usepackage{graphicx}
\usepackage{textcomp}
\usepackage{xcolor}
\usepackage{hyperref}
\usepackage{cite}
\usepackage[super]{nth}
\usepackage{makecell}

\usepackage{svg}

\usepackage{subfigure,url,float,bm,multirow,booktabs}

\usepackage{tabularx}

\usepackage{lipsum}

\DeclareMathOperator*{\argmax}{arg\,max}

\begin{document}

\title{RONELDv2: A faster, improved lane tracking method}

\author{Zhe Ming Chng, 
Joseph Mun Hung Lew, Jimmy Addison Lee
%\author{Yuguang ``Michael'' Fang,~\IEEEmembership{Fellow,~IEEE}

\thanks{Z. M. Chng is currently with the College of Computing, Georgia Institute of Technology, GA, USA. Email: zchng3@gatech.edu}% <-this % stops a space
\thanks{J. M. H. Lew and J. A. Lee are currently with Aison Pte. Ltd., Singapore. Email: \{joseph.lew, jimmy.lee\}@aison.sg}
%\thanks{Manuscript received XXX, XX, 2021; revised XXX, XX, 2021.}
}

%\markboth{IEEE Transactions on Intelligent Vehicles,~Vol.~XX, No.~XX, XXX~2021}
{}
%{Shell \MakeLowercase{\textit{et al.}}: Bare Demo of IEEEtran.cls for Journals}

\maketitle

\begin{abstract}
Lane detection is an integral part of control systems in autonomous vehicles and lane departure warning systems as lanes are a key component of the operating environment for road vehicles. In a previous paper, a robust neural network output enhancement for active lane detection (RONELD) method augmenting deep learning lane detection models to improve active, or ego, lane accuracy performance was presented. This paper extends the work by further investigating the lane tracking methods used to increase robustness of the method to lane changes and different lane dimensions (\textit{e.g.} lane marking thickness) and proposes an improved, lighter weight lane detection method, RONELDv2. It improves on the previous RONELD method by detecting the lane point variance, merging lanes to find a more accurate set of lane parameters, and using an exponential moving average method to calculate more robust lane weights. Experiments using the proposed improvements show a consistent increase in lane detection accuracy results across different datasets and deep learning models, as well as a decrease in computational complexity observed via an up to two-fold decrease in runtime, which enhances its suitability for real-time use on autonomous vehicles and lane departure warning systems.
\end{abstract}


\IEEEpeerreviewmaketitle

\section{Introduction}

Lane detection research has seen increased interest in recent years due to its importance in autonomous vehicles and advanced driver assistance systems (ADAS). Lane markings, in particular active or ego lane markings, serve as important markers that indicate the road lane the vehicle is currently travelling on and can be used to constrain the maneuver of the vehicle on public roads. This prevents collisions with other road users and provides the vehicle with a better understanding of its surroundings. 

To detect these lane markings, there has been much research on using different methods such as using cameras and light detection and ranging (LiDAR) sensors~\cite{Li2014}. Camera-based lane detection has been a popular method due to the disadvantages of other lane detection methods, such as the high cost of LiDAR sensors~\cite{Xing2018}. It also more closely mimics how humans detect lanes using visual cues and has seen increasing performance with the use of deep learning lane detection models, which further enhance its attractiveness for lane detection on vehicles. However, despite these advantages and recent improvements, vision-based lane detection faces significant challenges due to complicated driving scenarios with heavy traffic that occludes lane markings, poor weather conditions that blur important contours, as well as shadows and other road markings that could be easily mistaken for lane markings, \textit{etc}. However, lane detection also necessitates a real-time, lightweight solution that can be placed on edge devices in autonomous vehicles while being able to respond rapidly to changes in the dynamic and high-speed driving environment. While recent state-of-the-art lane detection methods have adopted new innovations, such as the attention mechanism~\cite{KYEOTL} and in particular the self-attention mechanism~\cite{LSTR,Hou2019}, to address some of these concerns, it is observed that many of these models still face difficulty adapting to datasets that differ significantly from their train sets. This covariate shift arising from the different distribution of train and test set data can be attributed to reasons such as varying road surface conditions and different lane markings. This is a cause for concern as autonomous vehicles must be equipped to respond to different driving environments and scenarios that might not have been presented in their train sets, hence a more robust method is required to improve lane detection results in such scenarios.

In our previous paper, we introduced the RONELD method~\cite{Chng2020} which aimed to improve lane detection results by enhancing the outputs of deep learning lane detection models through a four-part process involving adaptive lane point extraction, curved lane detection, lane construction, and lane tracking using preceding frames. In this paper, we further investigate the lane tracking step of the method and focus on observed rooms for improvement on the RONELD method's lane tracking step, such as the inability of RONELD to rapidly adapt to lane markings that appeared later on in the video feed if the original lane markings are still present in the frame and the constant distance relative to the frame width used to match lanes between frames which does not adjust for new lane marking thickness on different roads. We aim to further improve the ability of RONELD to obtain accurate detected active lanes, while remaining robust to changes in the road condition and driving environment. 

\begin{figure}[t]
	\centering
	\subfigure[Before]
	{
		\includegraphics[width=38mm]{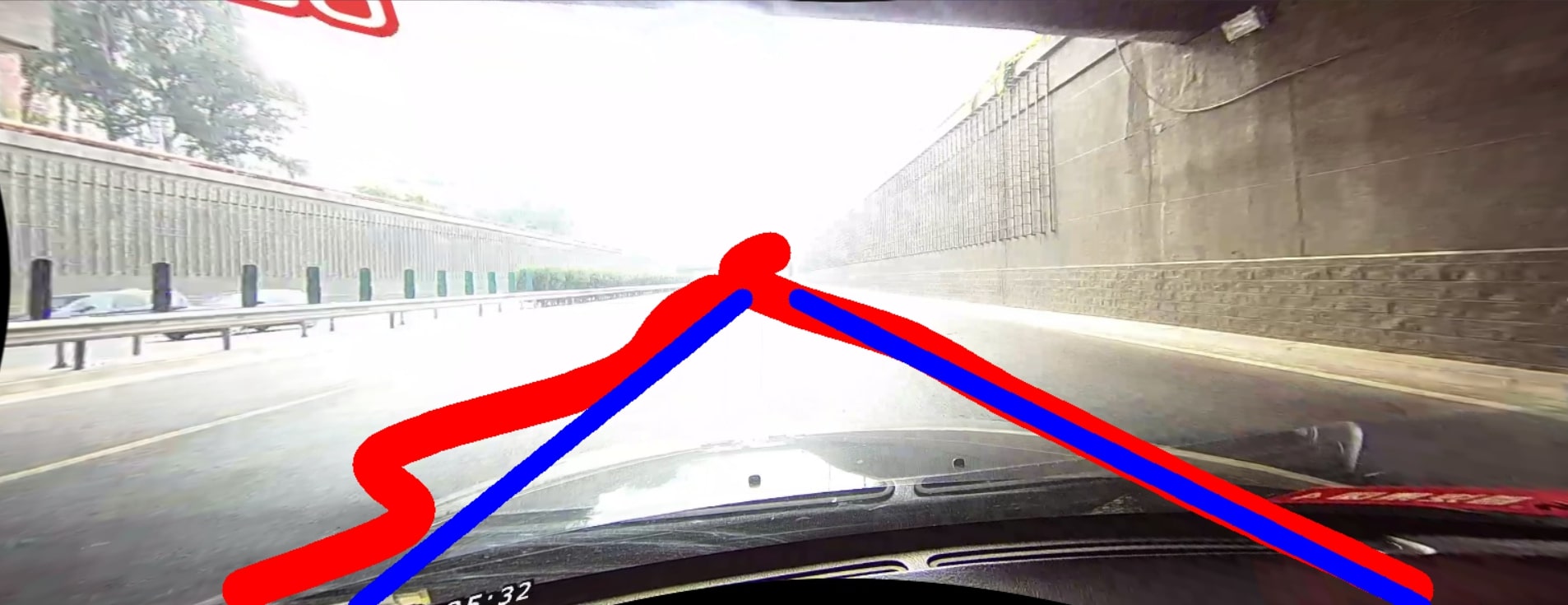}\label{fig1a}
	}
	\subfigure[After]
	{
		\includegraphics[width=38mm]{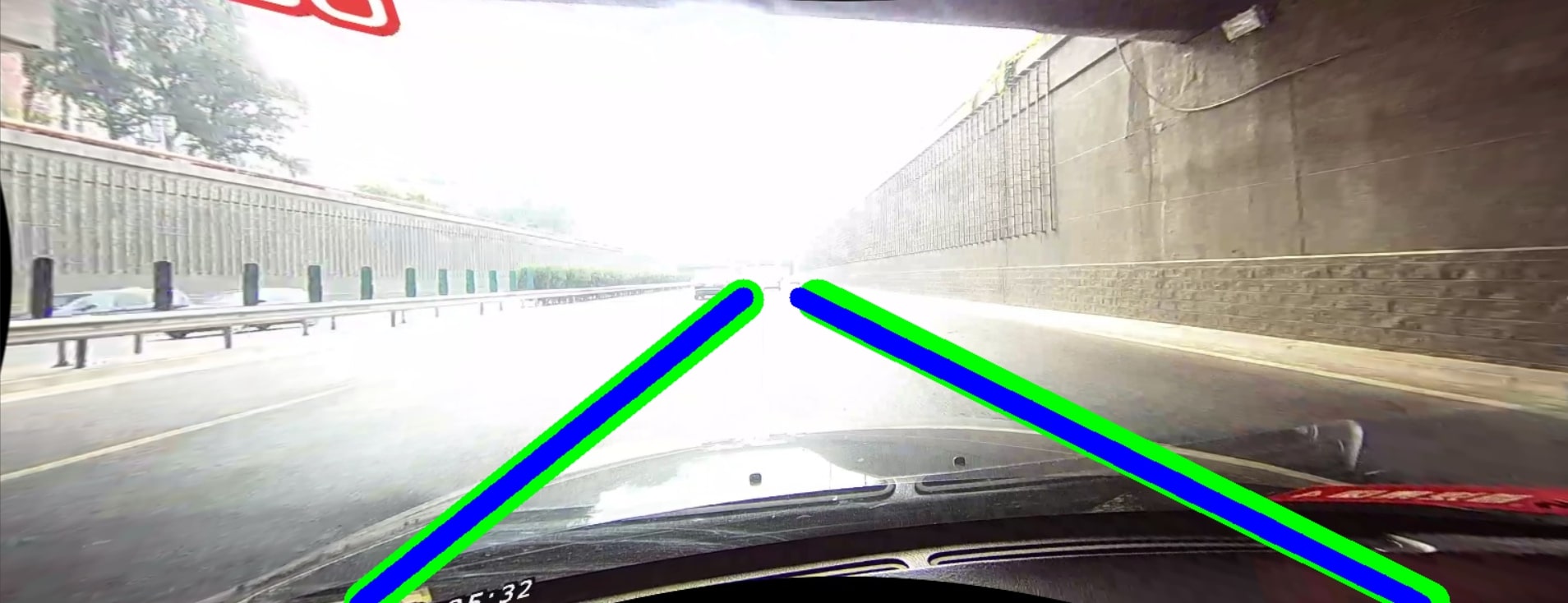}\label{fig1b}
	}
	\subfigure[Before]
	{
		\includegraphics[width=38mm]{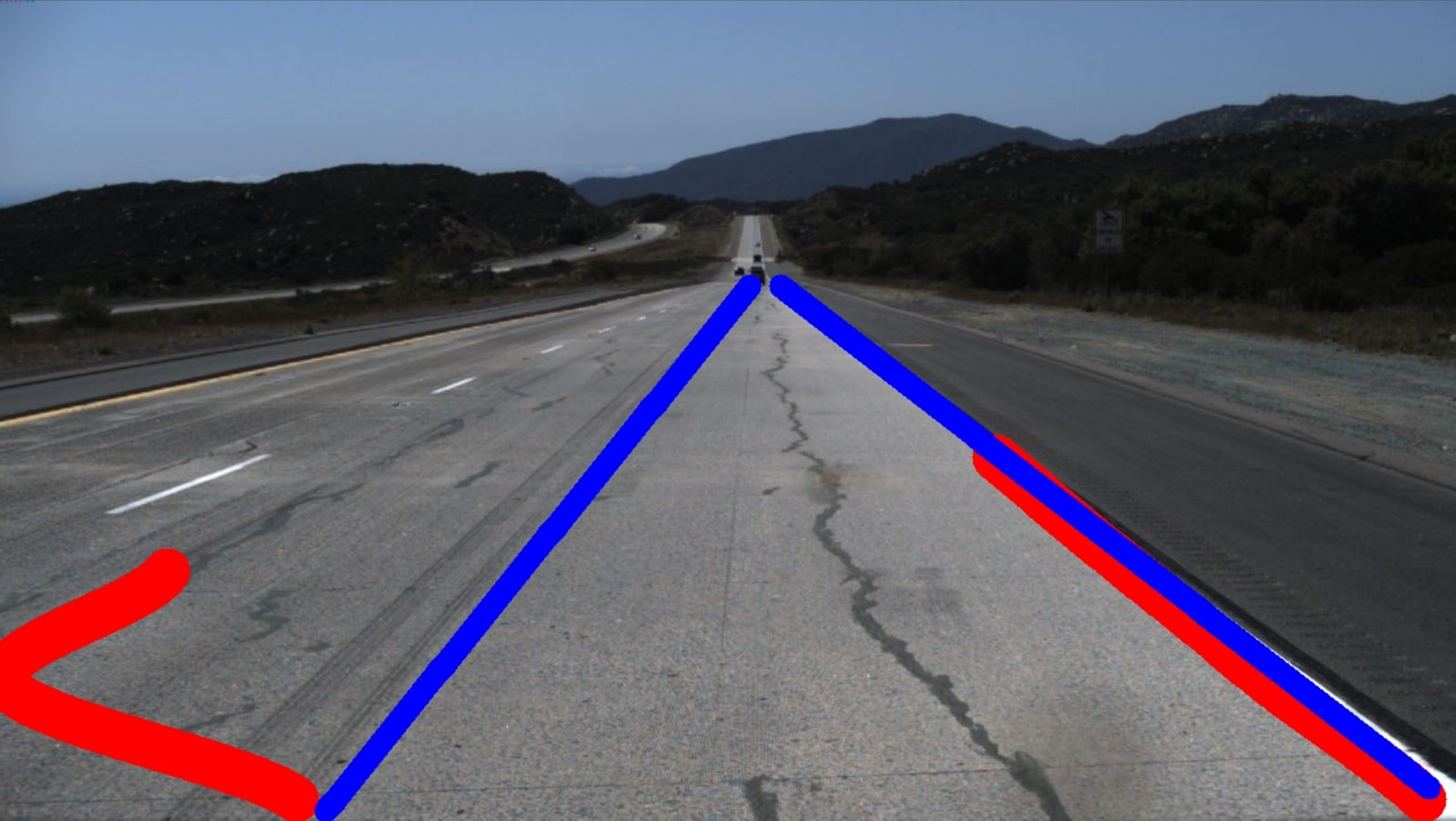}\label{fig1c}
	}
	\subfigure[After]
	{
		\includegraphics[width=38mm]{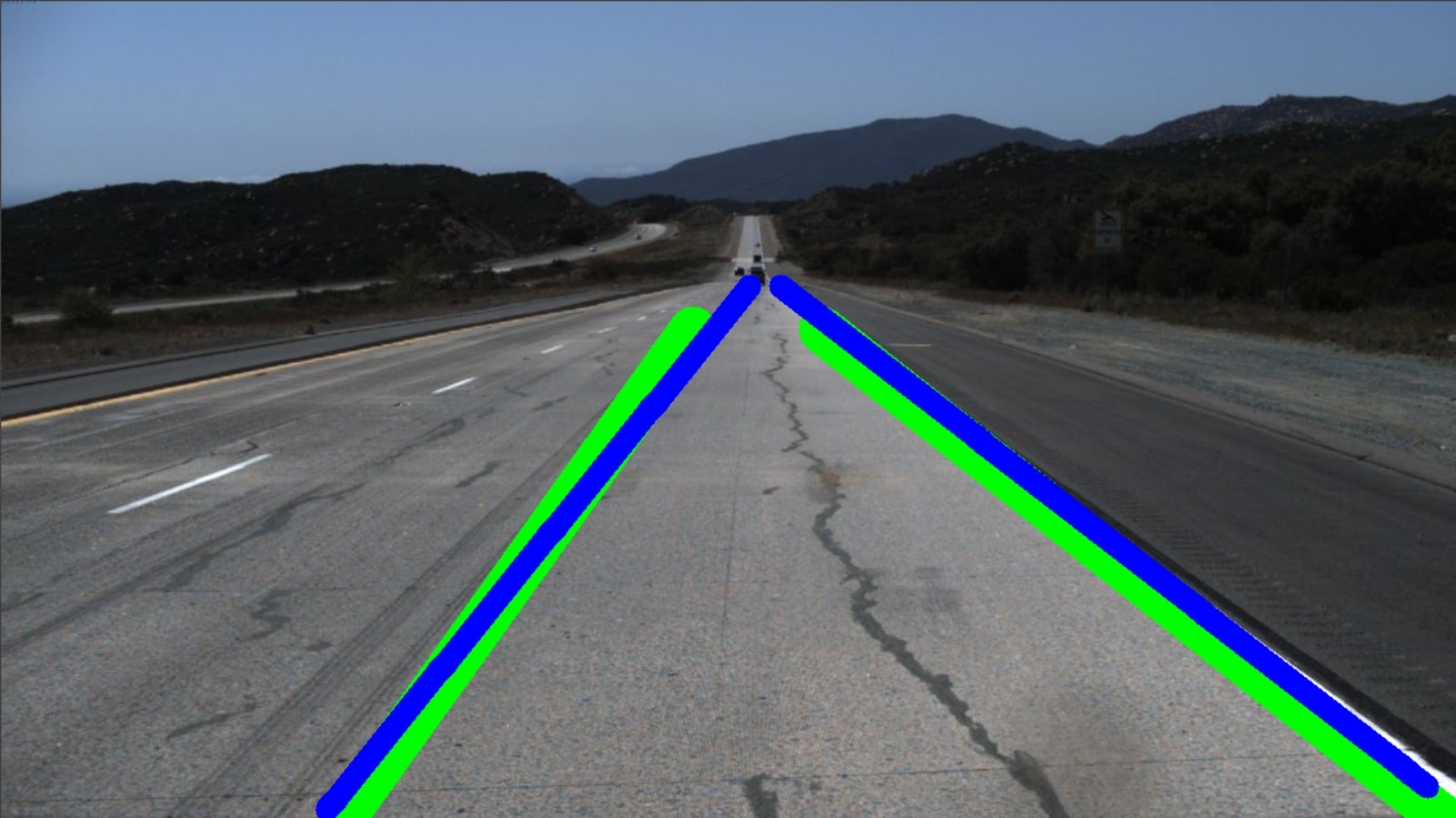}\label{fig1d}
	}
	\caption{Two SCNN lane detection results using a CULane-trained model on the CULane~\cite{Pan2018} and TuSimple~\cite{TuSimple-2019} test sets are shown in (a) and (c) respectively, detected lane marking in red, while ground truth lane markings are highlighted in blue. Corresponding results after RONELDv2 are shown in (b) and (d) respectively, highlighted in green.}\label{fig1}\end{figure}

To this end, we introduce RONELDv2, which incorporates the use of lane point variance, lane merging, and an exponentially weighted moving average method to compute weights in order to strive for a more robust and lightweight solution which further improves on the RONELD method and increases its suitability for real-time use on autonomous vehicles and ADAS. To verify the usefulness of our proposed changes to the method, we test the proposed changes on the TuSimple~\cite{TuSimple-2019} and CULane~\cite{Pan2018} datasets and compare accuracy results with those achieved in~\cite{Chng2020} as well as the two state-of-the-art models used in the RONELD paper, namely SCNN~\cite{Pan2018} and ENet-SAD~\cite{Hou2019}. Our experiments demonstrate the effectiveness of the changes, with increased accuracy results above those achieved by RONELD and an up to two-fold decrease in the runtime compared to an optimized version of RONELD. We present two simple before and after comparisons of applying RONELDv2 onto the CULane and TuSimple test set images in Fig.~\ref{fig1}.

We organize the rest of our paper as follows. Section II discusses related work on camera-based lane detection. Section III explains our proposed method in three parts: Lane point variance, weighing tracked lanes, and lane merging. Section IV provides empirical evidence of the usefulness of our method, and section V concludes our work.

\section{Related Work}

\subsection{Traditional lane detection}
A common process workflow for traditional lane detection algorithms utilizes a three-step process: pre-processing, lane detection, and lane tracking. To pre-process the images, multiple methods have been used such as edge detection algorithms (\textit{e.g.} Canny edge detector and Sobel operator), adaptive thresholding, and inverse perspective mapping~\cite{Wang2014}. For the lane detection step, there is a reliance on hand-crafted features, such as color-based features~\cite{Chiu2005}, steerable filters~\cite{McCall2004}, Hough transform~\cite{Li-2015,Haloi2015}, random sample consensus (RANSAC)~\cite{Borkar2009,Aly2008}, and ridge features~\cite{Beyeler2014,Lopez2010}. Many traditional lane detection algorithms apply some combination of these hand-crafted features in a lane detection step on the pre-processed road image, before applying a final tracking step. For the final tracking step, some algorithms utilize tracking algorithms, with particle~\cite{Loose2009} or Kalman~\cite{Borkar2009} filter being a popular choice in the literature, to reduce noise in the lane marking outputs and obtain more accurate lane estimates from the noisy input data. While these methods are able to achieve good performance on datasets with certain characteristics such as good weather, straight lanes, clear road markings, \textit{etc.}, they generally lack robustness and fail to perform well on images that do not follow underlying assumptions, which restricts possible driving environments and conditions out of the many diverse driving conditions that autonomous vehicles would have to encounter while in use. This could include road images with heavy traffic which obstructs a clear view of the lane marking or poor weather conditions, which significantly reduce the performance of these traditional lane detection methods.

\subsection{Deep learning lane detection}
Kim \textit{et al.} introduced a convolutional neural network (CNN) method combined with RANSAC to detect lanes in complicated road scenes~\cite{Kim2014}. Following that, a dual-view convolutional neural network (DVCNN) method~\cite{He-2016} was proposed by He \textit{et al.} which uses front-view and top-view images simultaneously to improve lane detection precision by eliminating false detections arising from moving vehicles, arrows, words, \textit{etc.}  Recently, end-to-end deep learning models~\cite{Qin2020,Li2020,Garnett2019} have seen an increase in popularity in lane detection research after achieving state-of-the-art results on other computer vision tasks~\cite{Redmon2016, resnet, Zou2020} and with greater availability of large-scale lane detection datasets~\cite{TuSimple-2019,Pan2018, Xu2020}. One common approach to address the lane detection problem has been treating it as a semantic segmentation task and using CNNs to formulate dense predictions for the road image, \textit{i.e.} predict whether each pixel in the road image is a part of a lane marking~\cite{Hou2019, Pan2018, He-2016}. Lee \textit{et al.} proposed a vanishing point guided network (VPGNet)~\cite{Lee2017}, a multi-task network looking at road and lane markings as well as the vanishing point to improve lane detection under adverse weather conditions. Later on, Pan \textit{et al.} proposed a Spatial CNN (SCNN)~\cite{Pan2018} method which enabled message passing between pixels across rows and columns of a layer which allowed the method to learn the strong spatial relationships amongst pixels across rows and columns of an image and won \nth{1} place in the TuSimple Benchmark Lane Detection Challenge. Subsequently, Hou \textit{et al.} proposed a self attention distillation (SAD)~\cite{Hou2019} method which showed compelling results when combined with the lightweight ENet~\cite{Paszke2016}, ResNet-18~\cite{resnet}, and ResNet-32~\cite{resnet} models, while running 10 times faster than SCNN for the SAD incorporated ENet model. More recently, Qin \textit{et al.}~\cite{Qin2020} introduced a method which treats the lane detection process as a row-based selection problem using global features, achieving compelling results with the ResNet-34 model on the CULane~\cite{Pan2018} dataset while running around 23 times faster than SCNN. Other methods have also approached the lane detection problem as an instance segmentation task~\cite{Nevan2018,Ko2020} or model lanes in three dimensions~\cite{Garnett2019}, the latter which has achieved competitive results even on the TuSimple dataset~\cite{TuSimple-2019}, an image-only lane detection dataset. Some Generative Adversarial Networks (GANs)~\cite{Goodfellow2014} have also been recently introduced to address the lane detection problem~\cite{liu2020, Ghafoorian2018}, where the network benefits from seeing both real and generated fake predictions at the same time to improve lane detection results.

\begin{figure}
    \centering
    \subfigure[Input]{
        \hspace{-2.5mm}
        \includegraphics[width=28mm,height=15.75mm]{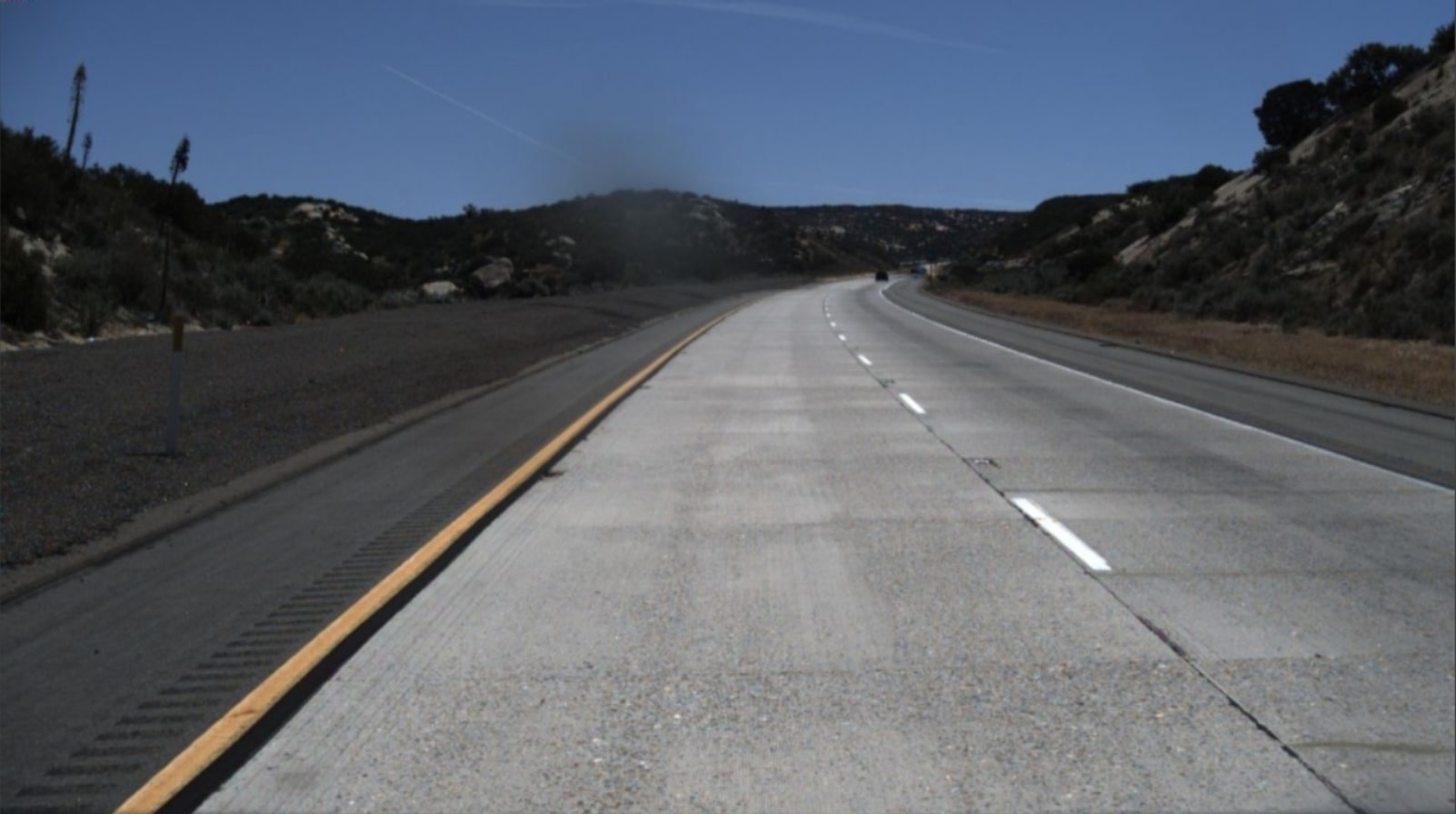}
        \hspace{-2.5mm}
    }
    \subfigure[SCNN]{
        \hspace{-2.5mm}
        \includegraphics[width=28mm,height=16mm]{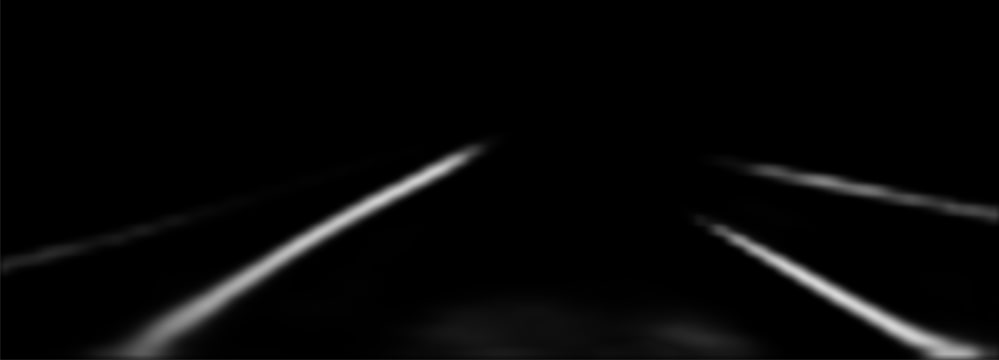}
        \hspace{-2.5mm}
    }
    \subfigure[ENet-SAD]{
        \hspace{-2.5mm}
        \includegraphics[width=28mm,height=16mm]{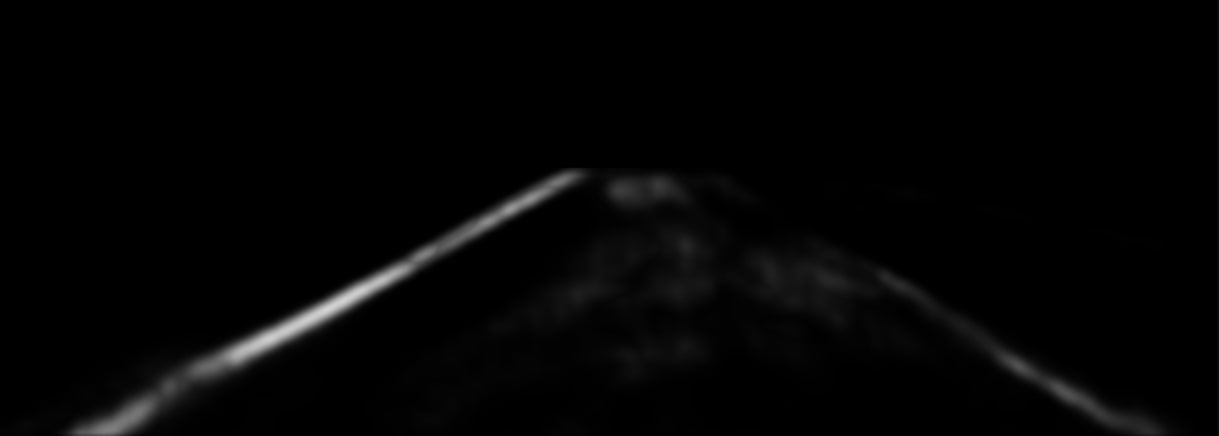}
        \hspace{-2.5mm}
    }
    \subfigure[Input]{
        \hspace{-2.5mm}
        \includegraphics[width=28mm,height=16mm]{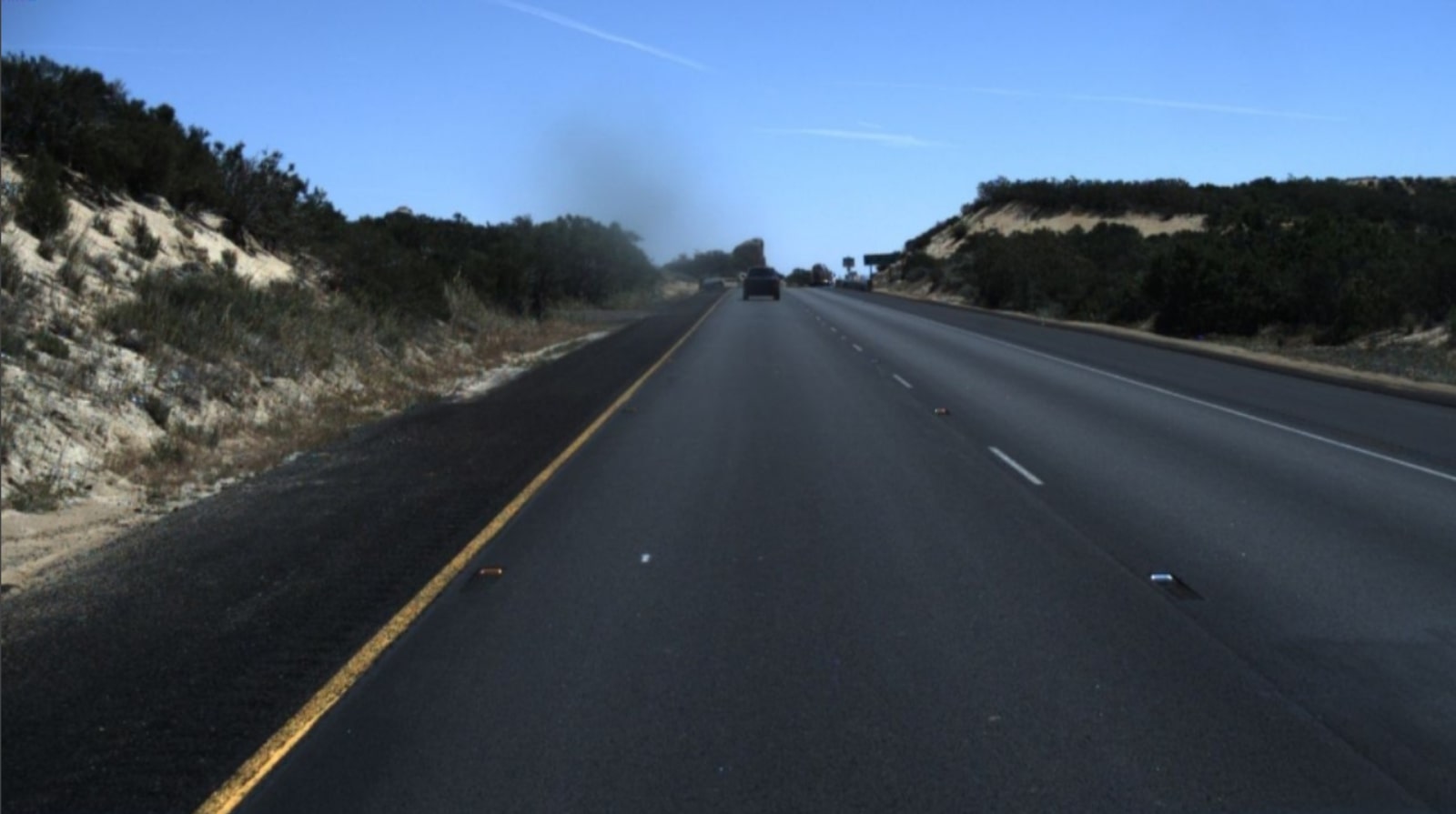}
        \hspace{-2.5mm}
    }
    \subfigure[SCNN]{
        \hspace{-2.5mm}
        \includegraphics[width=28mm,height=16mm]{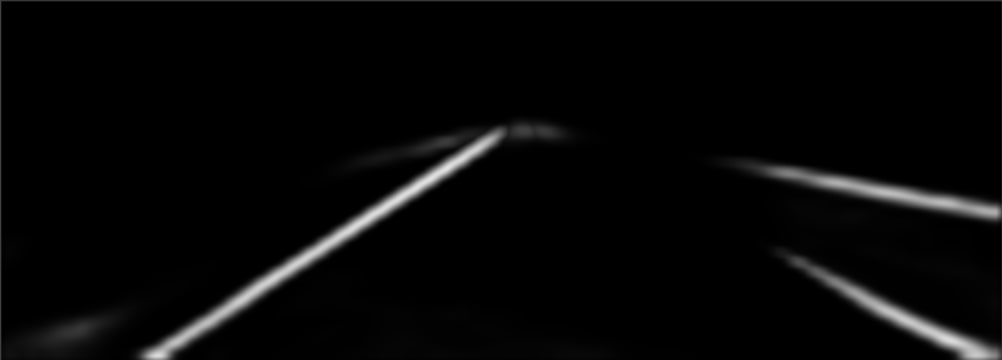}
        \hspace{-2.5mm}
    }
    \subfigure[ENet-SAD]{
        \hspace{-2.5mm}
        \includegraphics[width=28mm,height=16mm]{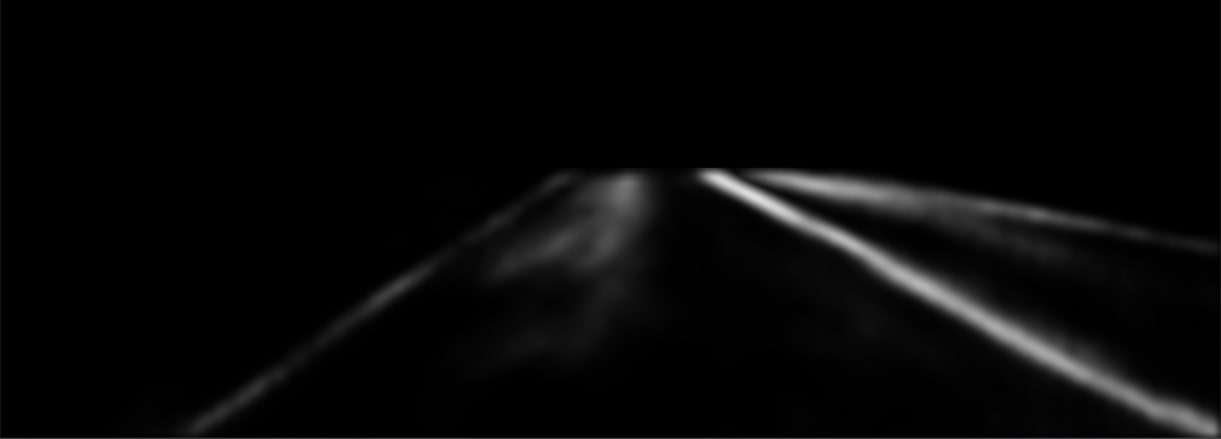}
        \hspace{-2.5mm}
    }
    \caption{Probability map outputs from the CULane-trained SCNN and ENet-SAD deep learning models on sample images from the TuSimple test set. }\label{proboutputfig}
\end{figure}

While the aforementioned methods have shown great promise in addressing the lane detection problem in a greater range of driving scenarios, they still face significant challenges adapting to lane images that are from a different distribution than their test set (\textit{e.g.} different traffic density, lane markings, or road surface conditions). To illustrate this issue, we have included probability maps from the cross-dataset validation of CULane-trained models on the TuSimple dataset in Fig.~\ref{proboutputfig}. There have been some methods using different techniques to enhance the performance of these models such as the use of an additional vanishing point prediction task added to guide VPGNet under adverse weather conditions, but they are usually paired for use with specific models or lack robustness.

\section{Proposed Method}

We explain our proposed method in this section. We build our contributions on top of the RONELD method, specifically focusing on improving the lane tracking portion of the method by making better use of information inherent in lane detection outputs from the deep learning models in previous frames.

\subsection{Lane point variance}\label{variancesection}

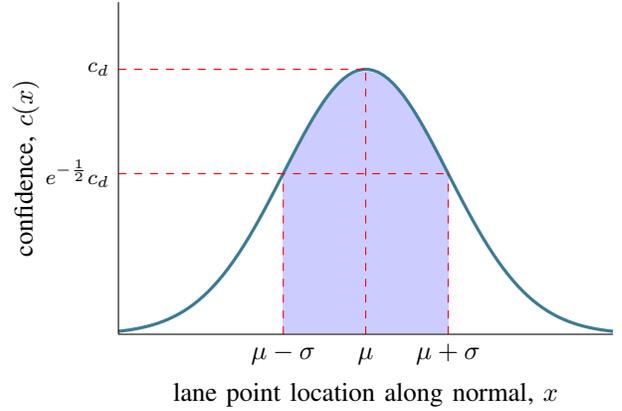
\begin{figure}
\begin{tikzpicture}
\begin{axis}[
  no markers, 
  domain=0:6, 
  samples=100,
  ymin=0,
  ymax=0.5,
  axis lines*=left, ylabel=\text{confidence, $c(x)$}, xlabel=\text{lane point location along normal, $x$},
  height=6cm, 
  width=\columnwidth-0.7cm,
  xtick=\empty, 
  ytick=\empty,
  enlargelimits=false, 
  clip=false, 
  axis on top,
  grid = major,
  ]

\begin{scope}[yshift=-\pgflinewidth]
\clip (axis cs:2,0) rectangle (axis cs:4,0.5);
\addplot [draw=none,fill=,blue!20!white] {gauss(x, 3, 1)};
\end{scope}

\addplot [very thick,cyan!50!black] {gauss(x, 3, 1)};
\pgfmathsetmacro\valueB{gauss(1,1,1.65)}
\pgfmathsetmacro\valueA{gauss(1,1,1.65)}
\pgfmathsetmacro\valueC{gauss(1,1,1)}

\draw [dashed, red] (axis cs:0,\valueB) -- (axis cs:4, \valueB);
\draw [dashed, red] (axis cs:0, \valueC) -- (axis cs:3, \valueC);

%vertical lines
\draw [dashed, red] (axis cs:4,0) -- (axis cs:4,\valueB);
\draw [dashed, red] (axis cs:2,0) -- (axis cs:2,\valueA);
\draw [dashed, red]  (axis cs:3,0) -- (axis cs:3,\valueC);

\node[left] at (axis cs:0,\valueC) {\footnotesize{$c_d$}};
\node[left] at (axis cs:0,\valueB) {\footnotesize{$e^{-\frac{1}{2}}c_d$}};

\node[below] at (axis cs:2,0) {$\mu-\sigma$};
\node[below, yshift=-0.57mm] at (axis cs:3,0) {$\mu$}; 
\node[below] at (axis cs:4,0) {$\mu+\sigma$};
\end{axis}
\end{tikzpicture}
\caption{Gaussian distribution graph with mean $\mu$ and variance $\sigma^2$, where the detected lane point is at the highest confidence point where $x=\mu$}\label{variancegraph}
\end{figure}

To enhance robustness to different lane marking thickness and different abilities of lane detection models to localize the lane marking, our method calculates the standard deviation of lane point positions from the output probability map of the deep learning model. To do this, we model the confidence values around the detected lane point as a normal distribution populated by points lying along the normal to the gradient of the proposed lane marking based on lane points detected by the RONELD method. This is achieved by using 
\begin{equation}
    c(x) = \frac{1}{\sigma\sqrt{2\pi}}e^{-\frac{1}{2}(\frac{x-\mu}{\sigma})^2},
\end{equation}
where $c(x)$ is the confidence of the point at $x$ based on Fig.~\ref{variancegraph}, $\mu$ is the mean of the distribution which in this case is the position of the detected lane point, and $\sigma$ is the standard deviation of the lane point position which we are searching for. These confidence markings are derived from the semantic segmentation output of the lane detection models. The detected lane point, which is the highest confidence point in the search area from the RONELD method~\cite{Chng2020}, occurs at $x = \hat{\mu}$, where $\hat{\mu} = \argmax_x c(x)$. If another point along the normal is found with a higher confidence, we would use that point as the detected lane point instead. Based on this model, $c(x) = e^{-\frac{1}{2}}c_d$ when $x = \mu\, \pm\, \sigma$, which our method uses to find $\sigma$ by finding the shortest distance to a point with confidence $c(x) \leq e^{-\frac{1}{2}}c_d$, where $c_d$ is the confidence of the relevant detected lane point, along the normal to the gradient of the lane. We do this for both the left and right side of the point and take the average of the two distances as the standard deviation $\sigma$ of that point, with the lane point variance as $\sigma^2$. We use the root mean square (RMS) of the standard deviation of the lane points in a lane to determine the standard deviation for the lane as a whole, which we use subsequently for our method.

%x, mean, variance
\pgfmathdeclarefunction{gauss}{3}{%
\pgfmathparse{1/(#3*sqrt(2*pi))*exp(-((#1-#2)^2)/(2*#3^2))}%
}

\begin{figure}
    \centering
    \subfigure[]{
        \includegraphics[height=30mm]{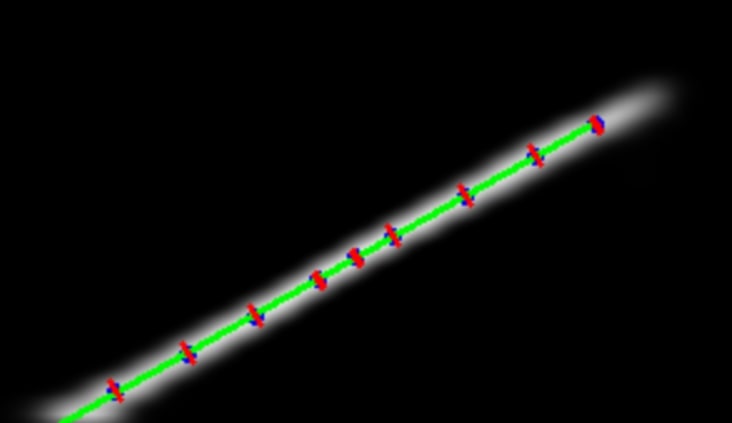}
    }
    \subfigure[]{
        \includegraphics[height=30mm]{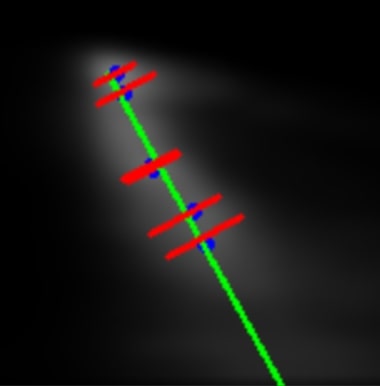}
    }
    \caption{Comparison of a (a) well-detected and defined lane marking and (b) less well-detected and defined lane marking. The blue points represent the detected lane points, the green line represents the weighted ordinary least squares regression line for the points and the red line indicates the normal to the regression line between the two points, $\mu - \sigma$ and $\mu + \sigma$, used in the calculation of $\sigma$ for each detected lane point.}
    \label{varianceimage}
\end{figure}

We use this when determining if a lane from a previous frame and a lane in the current frame are the same lane. Similar to RONELD, we calculate the RMS distance between the lanes. However, unlike RONELD which uses a fixed distance threshold based on image width to determine if lanes in previous and current frames should be matched, we set the threshold based on the standard deviation of the lane points instead, which allows our method to adjust for thicker lane markings that likely have greater variance in their lane points since more points fall within the lane width, or lanes that are detected with greater positional uncertainty by the deep learning model as seen in Fig. \ref{varianceimage}. This is critical as the lane points of these lanes are likely to see greater shift in their proposed location from frame to frame, leading to greater uncertainty in the lane parameters and hence greater distance despite being the same lane.

We also include the lane point variance information in the weights matrix for weighted ordinary least squares linear regression used in RONELD to determine the gradient and y-intercept for straight lanes, 

\begin{equation}
    \bm{C} = \begin{pmatrix}
    \frac{c_1}{{\sigma_1}^2} & 0 & \dots & 0\\
    0 & \frac{c_2}{{\sigma_2}^2} & \dots & 0\\
    \vdots & \vdots & \ddots & \vdots\\
    0 & 0 & \dots & \frac{c_m}{{\sigma_m}^2}
    \end{pmatrix},
\end{equation}
where $c_i$ is the confidence of the $i$-th detected lane point, and $m$ is the number of detected lane points. This is in contrast to the weights matrix used in RONELD which consisted solely of the confidence of the lane points as their respective weights. By taking into account the variance of the lane points in addition to the confidence values when computing the weight matrix for the weighted ordinary least squares linear regression step, our method adjusts for points that have greater positional uncertainty, represented by a higher lane point variance and lower confidence values. This further reduces the problem of heteroskedasticity due to differences in variance of lane points arising from factors such as the accuracy of the model on the dataset and different lane marking thickness.

\subsection{Weighing tracked lanes}

After matching the two lanes together based on the line distance with a threshold based on $\sigma$, we assign weights to each lane and the lanes with the highest weight in the left and right half of the frame are identified as the final lane marking outputs. In RONELDv2, we propose an exponentially weighted moving average (EWMA) method to replace RONELD's linear weight increment method. The benefits of this new method are two-fold. Firstly, it increases the robustness of the system to new lanes by limiting the relative weight of old lanes compared to new lanes, where old lanes are lanes that have appeared in many consecutive frames while new lanes are lanes that have appeared in comparatively fewer frames. Secondly, by reducing the ratio of lane weights between old and new lanes, it allows for old lanes that are not detected in the current frame to be cleared from the cache of previous lanes more rapidly, reducing processing times needed to process the cache of previous lanes.

\begin{figure}
    \centering
    \subfigure[]{
        \includegraphics[width=0.46\columnwidth]{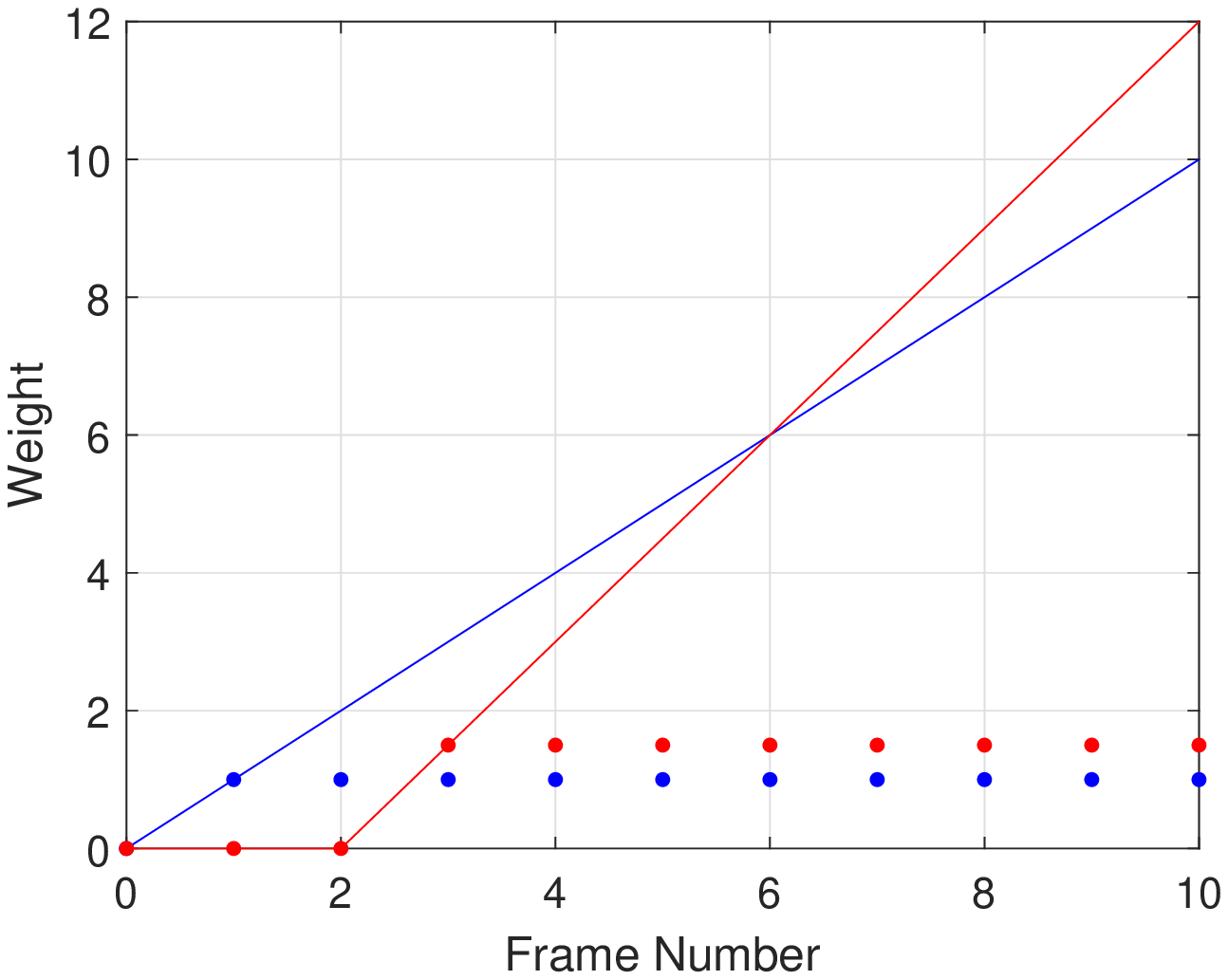}\label{lineargraph}
    }
    \subfigure[]{
        \includegraphics[width=0.46\columnwidth]{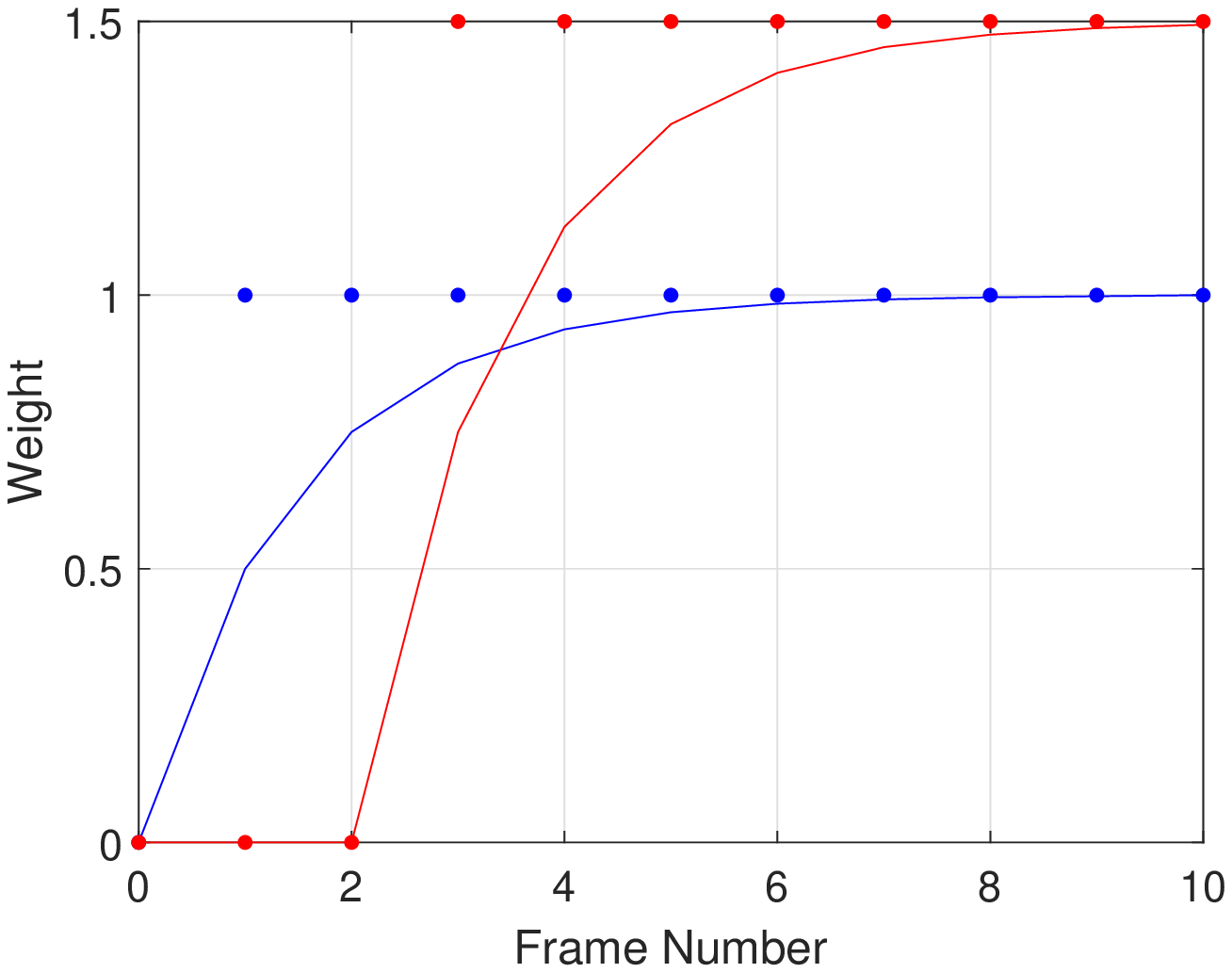}\label{exponentialgraph}
    }
    \caption{Comparison of a \protect\subref{lineargraph} linear weights method used in~\cite{Chng2020} against an \protect\subref{exponentialgraph} exponentially weighted moving average method ($\alpha=0.5$) used in our method. The red line represents the weight of a higher-weighted lane that appears from frame 3 onwards while the blue line represents the weight of a lower-weighted lane that appears from frame 1. The dots represent the weight of each lane in each frame.}
    \label{weightsfig}
\end{figure}

We calculate a current weight, $\omega_f$, for the lane in the current frame $f$ using
\begin{equation}
    \omega_f = \psi\,c_f N_f,
\end{equation}
where $\psi$ is the weight increment factor which is higher for identified potential active lane markings and lower for nonactive lane markings, $c_f$ is the RMS confidence of detected lane points, and $N_f$ is the number of lane points in frame $f$. This is similar to the weight increment used in RONELD. Using the $\omega_f$ of the lane in frame $f$ and the EWMA weight of the lane in preceding frames, we then calculate the current EWMA lane weight, $\Omega_f$, which we obtain using
\begin{equation}
    \Omega_f = \alpha\, \omega_f + (1 - \alpha)\,\Omega_{f-1},
\end{equation}
where $\alpha$ is a smoothing coefficient that represents the weight placed on the current frame when computing the EWMA weight. We use $\Omega_f$ as the final lane weight when determining the final lane marking outputs. The function can be seen as a piecewise function changing at each integer $f$ value (with $\omega_f$ changing at each integer $f$ as well) and with the gradient at each point calculated as 
\begin{equation}
    \frac{d\,\Omega}{df} = \alpha (\omega_f - \Omega).
\end{equation}
We set $\alpha = 0.5$ in this paper to place equal weight between the current lane weight and the previous EWMA lane weight. We use the EWMA function without bias correction to provide a weight advantage for lanes that have appeared in comparatively more frames. By using an EWMA function, it reduces the total weight of the lanes and imposes max$(\omega_f)$ as an upper bound on the weight of the lane and thereby limiting the weight of lanes despite appearing in many preceding frames. This reduces the weight of old lanes more rapidly and allows new lanes to quickly supersede old lanes while simultaneously allowing us to remove the old lanes from the stored cache of previous lanes, thereby reducing the processing required to handle these lanes as well. Assuming a constant lane weight $\omega_f$ in each frame, this can also be modelled as a geometric sequence where
\begin{equation}
    %aw is the first term, the ratio is (1 - a)
    \Omega_f = \omega[1 - (1-\alpha)^n], \, \omega = \omega_f,
\end{equation}
where $n$ is the number of frames that lane has appeared in. The benefits of this EWMA function can also be seen through an example in Fig.~\ref{weightsfig} using lanes with constant weights. In the example, the higher-weighted lane with constant weight 1.5 appears in frame 3 and overtakes the lower-weighted lane with constant weight 1 more rapidly, in frame 4 instead of frame 6, while converging towards the weight of the lane in each frame. 

%if Omega approximately equals omega, the difference in gradient is always larger for exponentially weighted average but the degree of how much is determined by the alpha value, because exponential gradient diff - linear gradient diff = alpha(omega_f - Omega_f) - (omega_f - omega), where omega is the old lane's weight, equals (1 - alpha) omega_f - Omega_f + omega and we're looking at Omega_f < omega since that's the part where new lane has smaller age than old lane, so gradient always greater until new lane exceeds old lane

\begin{figure}
    \centering
    \includegraphics[height=45mm]{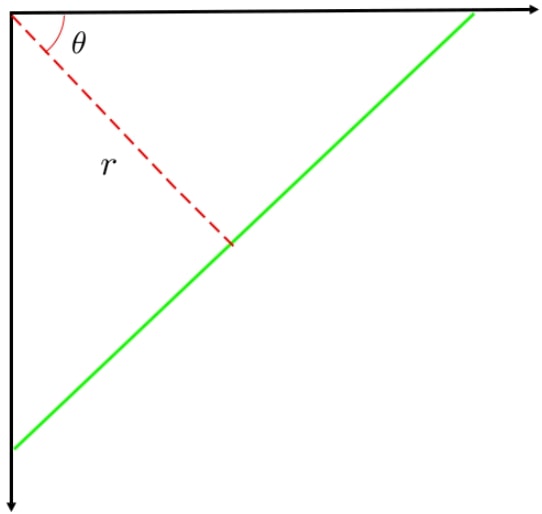}
    \caption{Illustration of ($r$, $\theta$) lane parameters used in lane merge step}\label{houghlineimage}
\end{figure}

%table of dataset description
\begin{table*}
	\small
	\caption{Basic information of datasets (TuSimple and CULane) used in our experiments.} \label{datasettable}
	\centering
	\vspace{-1mm}
	\begin{tabular}{| c | c c c c c c|}
		\hline
		Dataset
		& \# Total
		& \# Test 
		& \# Test Labelled
		& Resolution
		& Environment
		& Traffic Density\\
		\hline
		\hline
		TuSimple
		& 6,408
		& 55,640
		& 2,782
		& 1280$\times$720
		& Highway 
		& Low\\
		CULane
		& 133,235
		& 34,680
		& 34,680
		& 1640$\times$590
		& Urban, rural, highway 
		& Varies\\
		\hline
	\end{tabular}
    \vspace{-1mm}
\end{table*}

\subsection{Lane merging}\label{lanemergesection}

To further enhance the ability of our method to adjust to noise in the lane detection model's semantic segmentation output, we use lane merging to obtain current lane parameters from the current and previous frames. As a pre-processing step, we convert the lane parameters obtained from the weighted least squares linear regression to Hesse's normal form with the line parameters ($r$, $\theta$), where $r$ is the length of the perpendicular from the origin to the line and $\theta$ is the angle between that perpendicular and the $x$-axis. An illustration of the lane parameters is provided in Fig. \ref{houghlineimage}. We use this form as it prevents a line with a large gradient from dominating the lane parameters when merging the lane parameters together. As an example, a vertical line has a gradient tending to infinity, and hence the weighted average lane would adopt a gradient tending to infinity as well, regardless of the gradient of the other lane being measured. 

Using these lane parameters, we implement a lane merging method that calculates the lane parameters in the current frame by weighing the lane parameters from the previous and current frame with $\frac{\Omega_{f-1}}{\sigma_{f-1}}$ and $\frac{\omega_f}{\sigma_f}$ as the weights respectively. Using these weights, the proportion of the current lane parameters contributed by the lane in the current frame,  $\zeta_f$, is equal to 
\begin{equation}
    \zeta_f = \frac{\omega_f\,\sigma_{f-1}}{\omega_f \,\sigma_{f-1} + \Omega_{f-1}\,\sigma_f}.
\end{equation}
Using these weights, we can obtain the predicted lane parameters for the current frame, ($r_f$, $\theta_f$), from the detected lane parameters, ($\hat{r}_f$, $\hat{\theta}_f$) and lane parameters from previous frame, ($r_{f-1}$, $\theta_{f-1}$) through

\begin{equation}
    \begin{pmatrix}
    r_f\\[3pt]
    \theta_f
    \end{pmatrix} = \begin{pmatrix}
    \hat{r}_f & r_{f-1}\\[3pt]
    \hat{\theta}_f & \theta_{f-1}
    \end{pmatrix}
    \begin{pmatrix}
    \zeta_f\\[3pt]
    1 - \zeta_f    
    \end{pmatrix}
\end{equation}
which allows us to use lane information over multiple frames to deduce more accurate parameters for the lane in the current frame, similar to other estimation algorithms such as the Kalman filter but without prior assumptions on the distribution of process or observation noise, hence allowing the method to remain a turnkey solution that can be easily added to deep learning models for new datasets without fitting new parameters.

\section{Experiments}
To verify the usefulness of our proposed changes to RONELD and to check the accuracy results of our proposed method, we run experiments on the test sets of two popular lane detection datasets in the literature, namely TuSimple~\cite{TuSimple-2019} and CULane~\cite{Pan2018}, which were also used in RONELD~\cite{Chng2020}. 

\begin{figure}
	\centering
	
	\includegraphics[height=18mm, width=39.6mm]{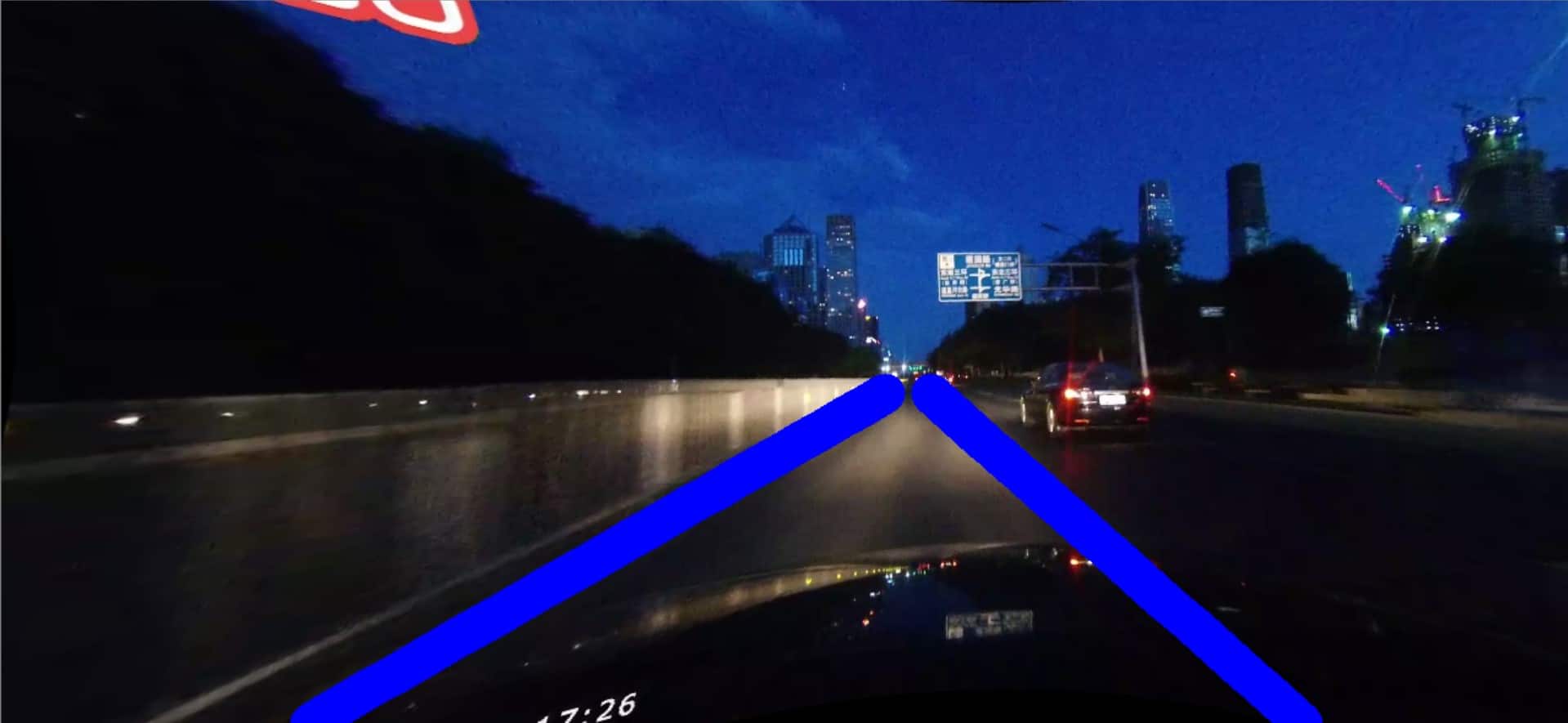}
	\hspace{1.5mm}
	\includegraphics[height=18mm, width=32mm]{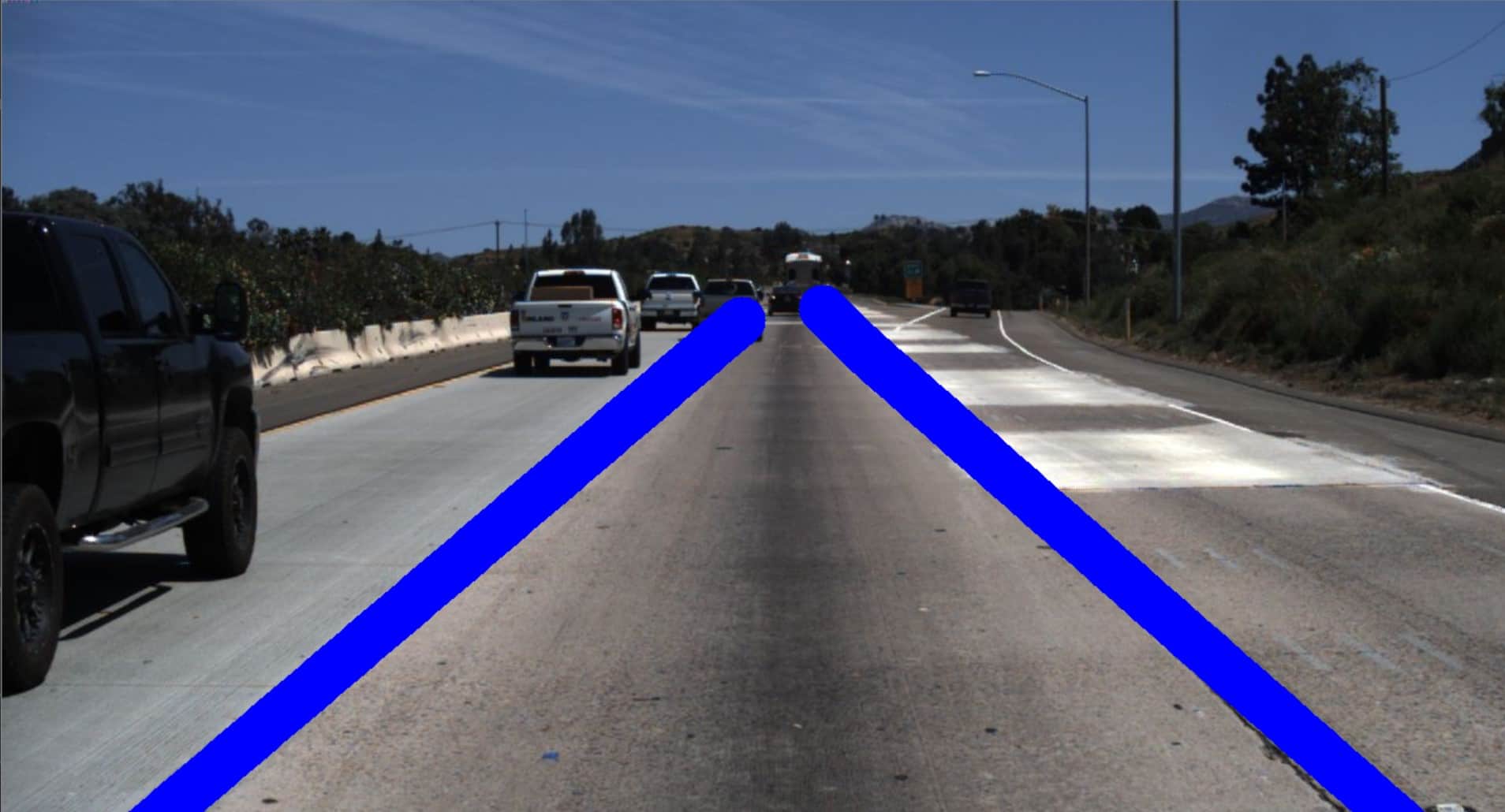}
	\vspace{2mm}
	
	\includegraphics[height=18mm, width=39.6mm]{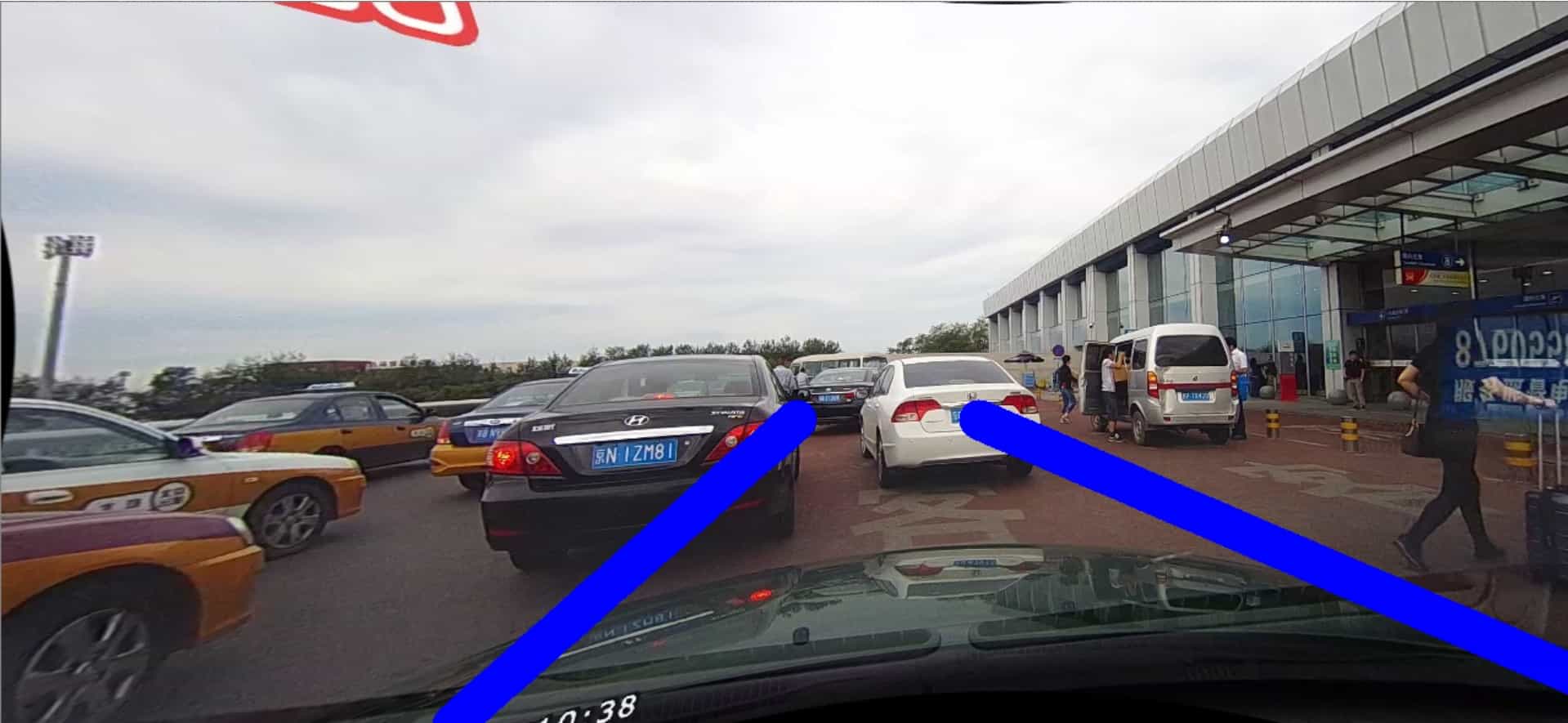}
	\hspace{1.5mm}
	\includegraphics[height=18mm, width=32mm]{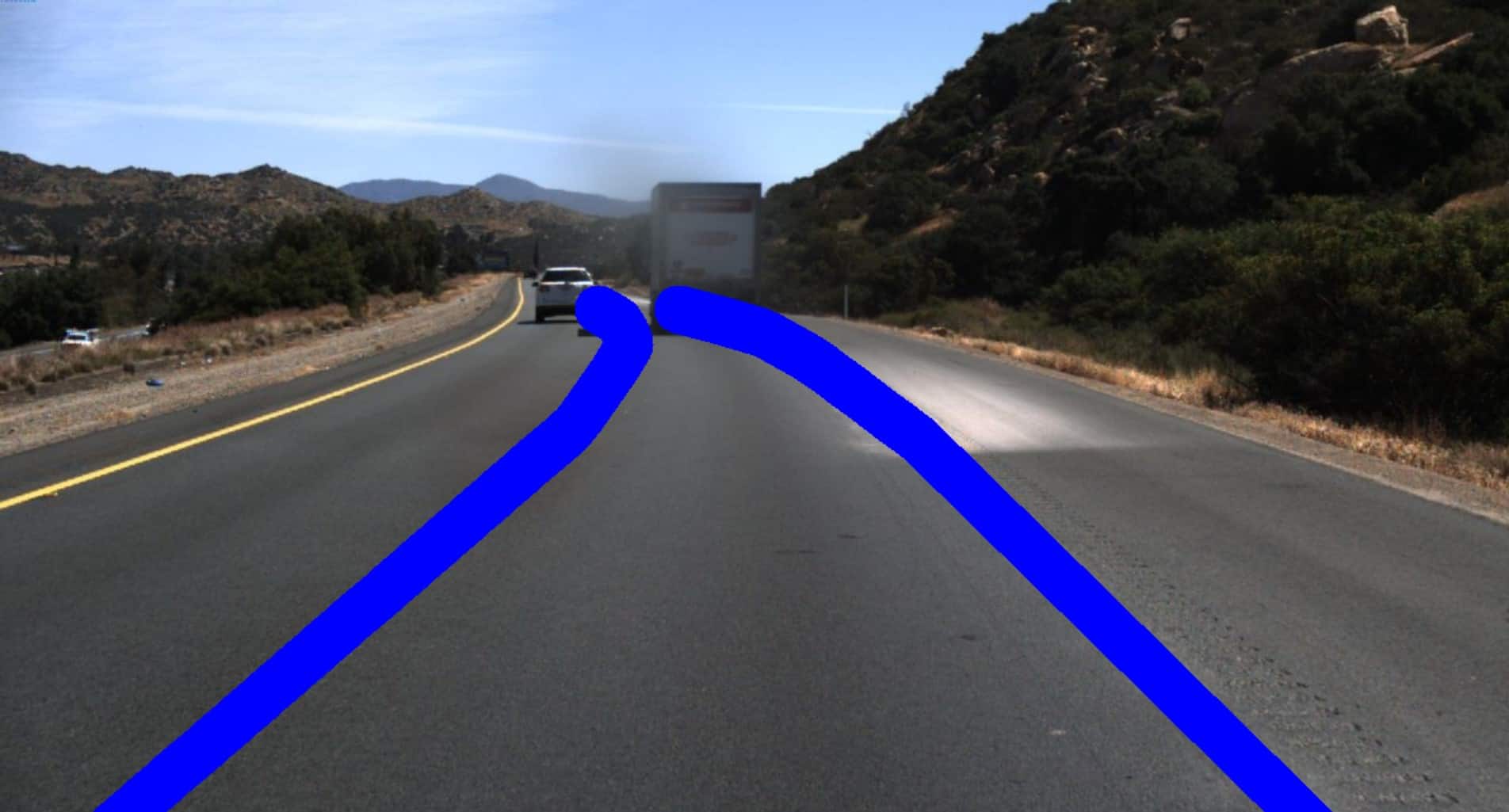}
	\vspace{2mm}
	
	\includegraphics[height=18mm, width=39.6mm]{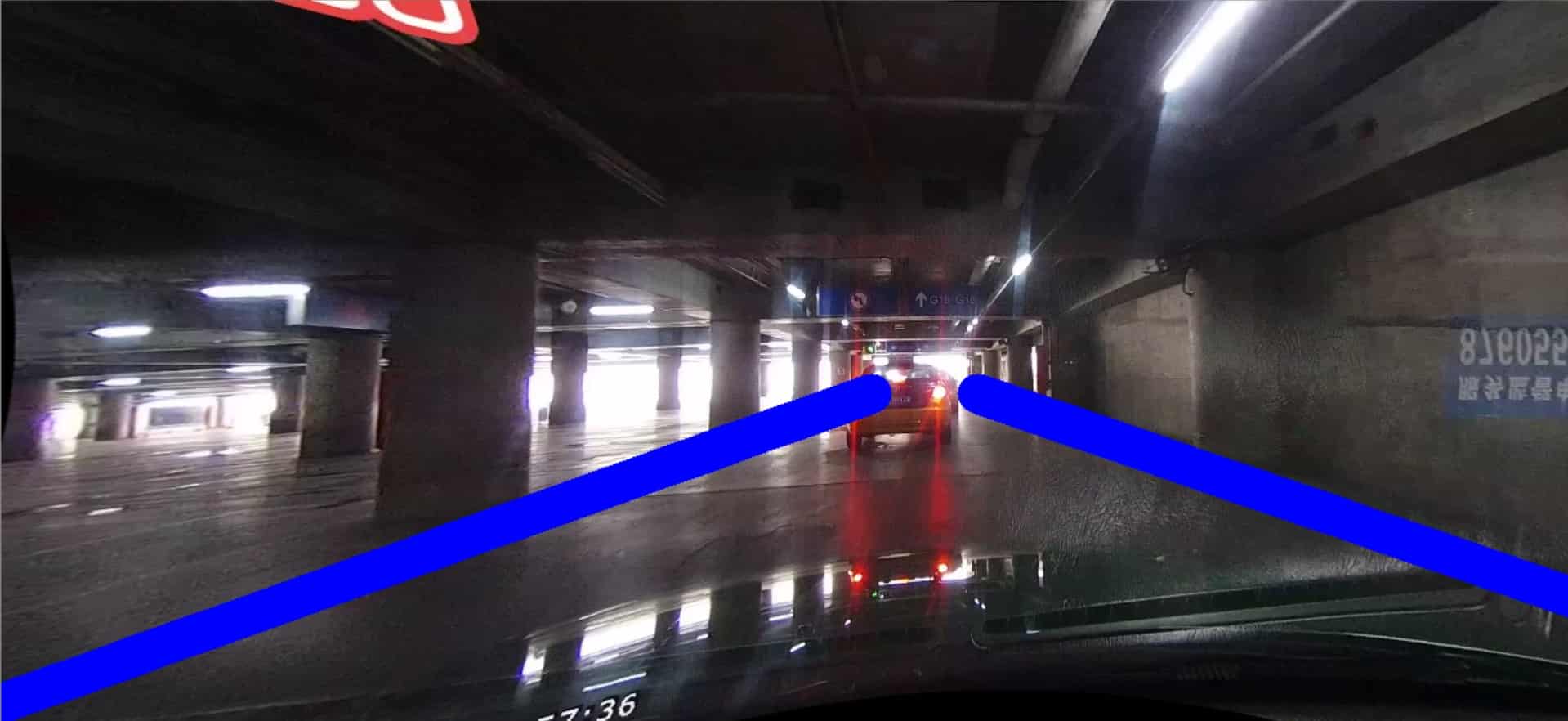}
	\hspace{1.5mm}
	\includegraphics[height=18mm, width=32mm]{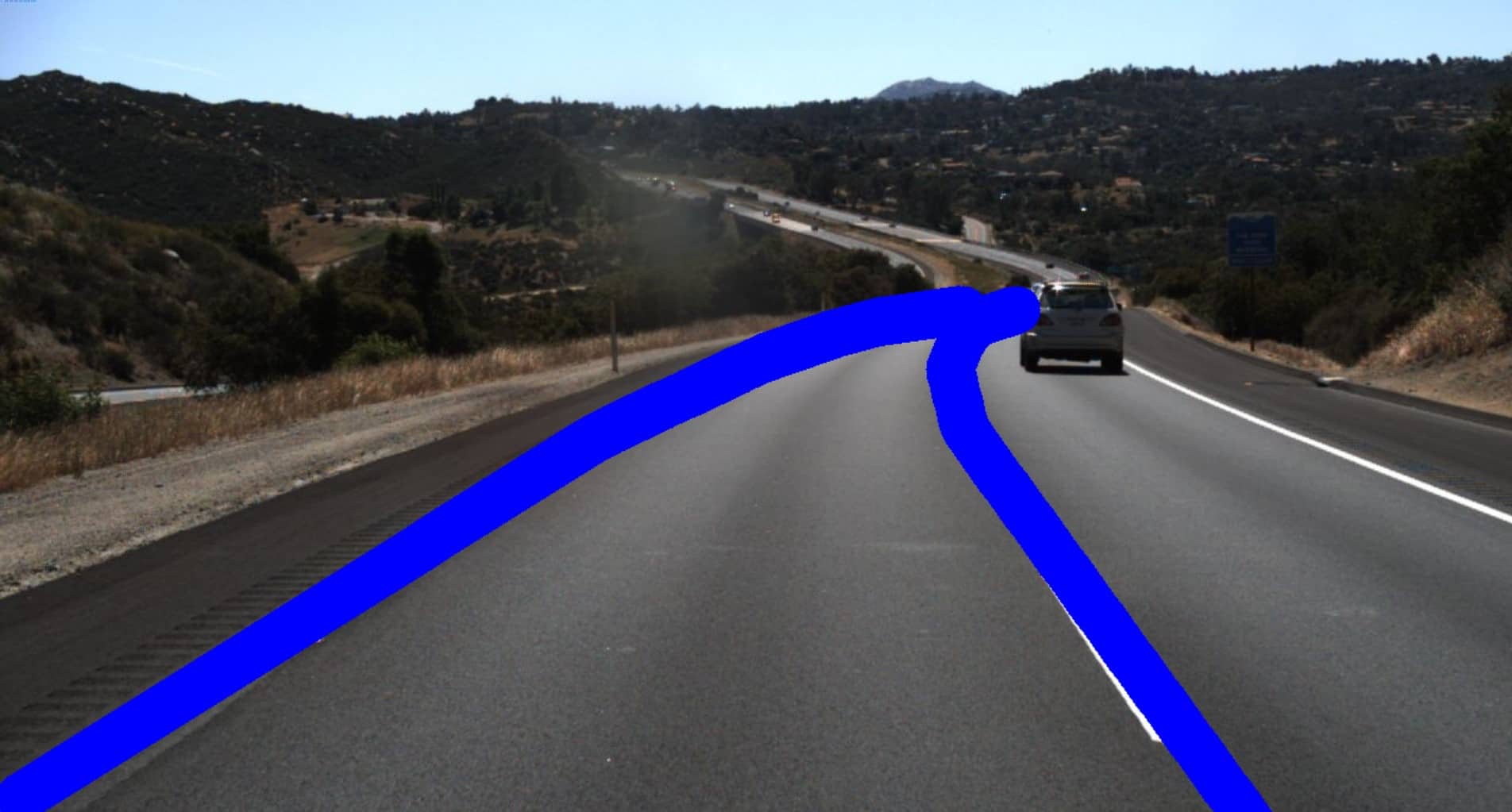}
	\vspace{0.4mm}
    
	\subfigure[]{
	    \hspace{-1.3mm}
		\includegraphics[height=18mm, width=39.6mm]{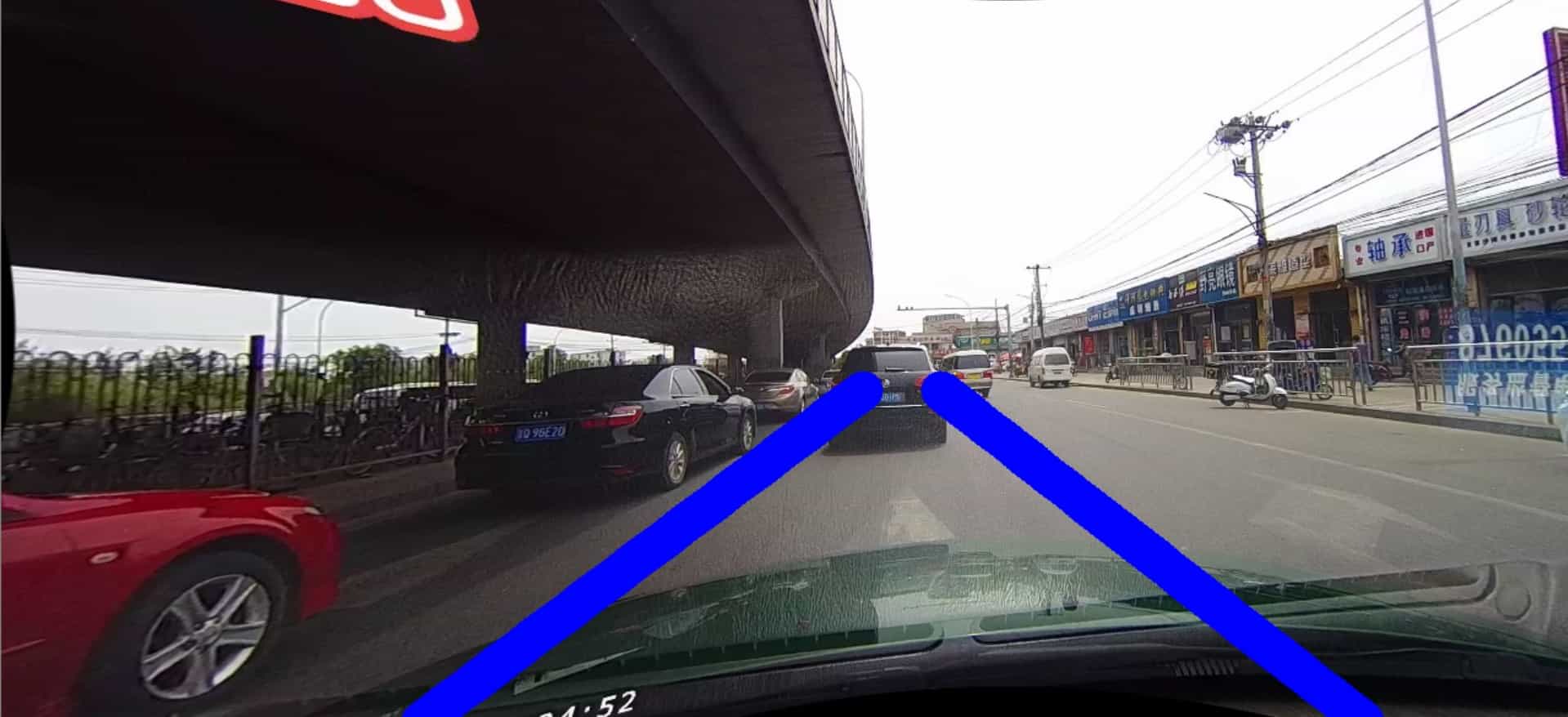}
		\,
	}
	\hspace{-2mm}
	\subfigure[]{
	    \hspace{-2mm}
		\includegraphics[height=18mm, width=32mm]{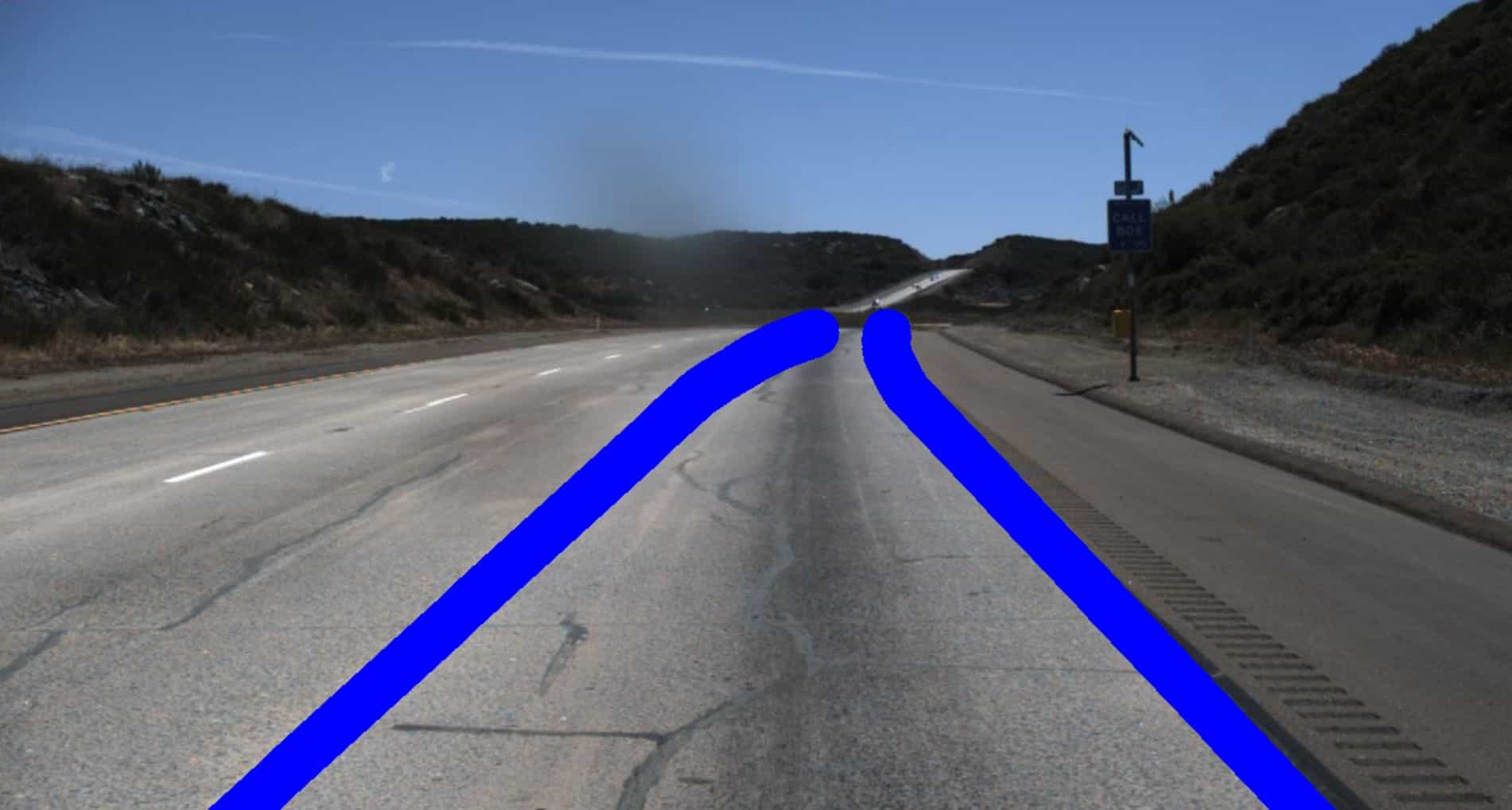}
	}
    \vspace{-1mm}
	\caption{Sample images from the (a) CULane and (b) TuSimple test sets. Ground truth active lane markings are highlighted in blue. }\label{samplepics}\end{figure}

\subsection{Dataset details}
Table~\ref{datasettable} summarizes the details of the two datasets and Fig.~\ref{samplepics} shows some sample frames from the datasets. Comparing the datasets, CULane is generally a more complicated dataset with a variety of road conditions and traffic densities, which present challenges due to wider variety of road image contexts and more challenging driving scenarios (\textit{e.g.} night scenes with poor lighting conditions and occlusions of lane markings by other vehicles). In contrast, TuSimple is a rather simple dataset consisting primarily of highway roads in fair weather conditions with low traffic density, albeit containing some variability in road surface color and conditions. TuSimple also has ground truths labelled on the last frame of each twenty-frame clip, which provides preceding frames for our lane tracking method and RONELD to use before making predictions on a labelled image. On the other hand, CULane provides ground truths on all road images within the dataset.

\subsection{Evaluation metric}
To identify true positive (TP) predicted lane markings, we use the intersection over union (IoU) between ground truth and predicted lane markings, similar to~\cite{Hou2019,Pan2018,Chng2020}. This method sets the line width of the ground truth lane markings and predicted lane markings as 16 and 30 pixels respectively. We use these line widths for output probability maps with widths of 800 pixels and scale them accordingly for different output probability map sizes for uniform comparison across models. Lane predictions with IoU values above a certain threshold are marked as TP lanes for that threshold. We then calculate the accuracy of each model for each threshold value using
\begin{equation}
    accuracy = \frac{N_{TP}}{N_{gt}},
\end{equation}
where ${N_{TP}}$ is the number of TP lanes for a particular IoU threshold value and $N_{gt}$ is the number of ground truth lane markings. We record the accuracy at IoU threshold values between 0.3 and 0.5 (inclusive) at 0.1 intervals. We apply this evaluation metric consistently across our experiments for a uniform comparison across the original dataset and cross-dataset validation tests.

\subsection{Implementation details}
For our experiments we exploit two state-of-the-art lane detection models, SCNN~\cite{Pan2018} and ENet-SAD~\cite{Hou2019}. We use their probability map outputs with our method for comparison with the RONELD~\cite{Chng2020} method and the original models. We used models that are pre-trained on the CULane dataset, with no images from the TuSimple dataset as we use the TuSimple dataset for our cross-dataset validation tests to determine the performance of the methods on unseen datasets. We select the CULane model as the train dataset and TuSimple as the cross-dataset as CULane is a more complicated dataset and should ideally be generalizable to the TuSimple test set given its greater complexity and wider variety of driving scenarios.

\begin{figure}
    \centering
    \subfigure[Accuracy results on CULane test set using CULane-trained models.]{
        \hspace{-4.5mm}
        \includegraphics[width=0.90\columnwidth]{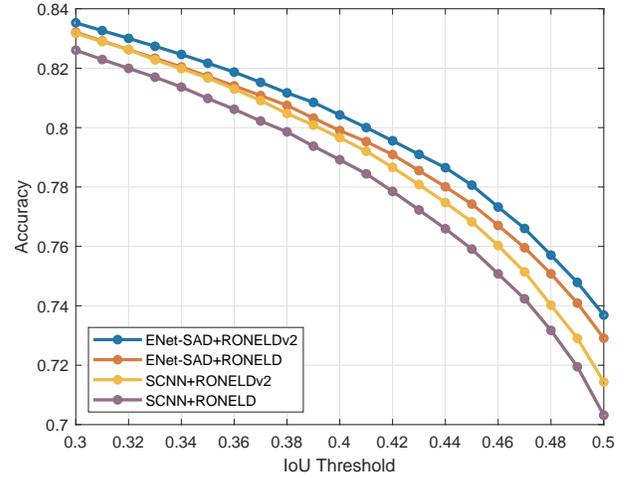}\label{culane_graph}
    }
    \vspace{4mm}
    \subfigure[Accuracy results on TuSimple test set, which is a cross-dataset validation using CULane-trained models.]{
        \includegraphics[width=0.89\columnwidth]{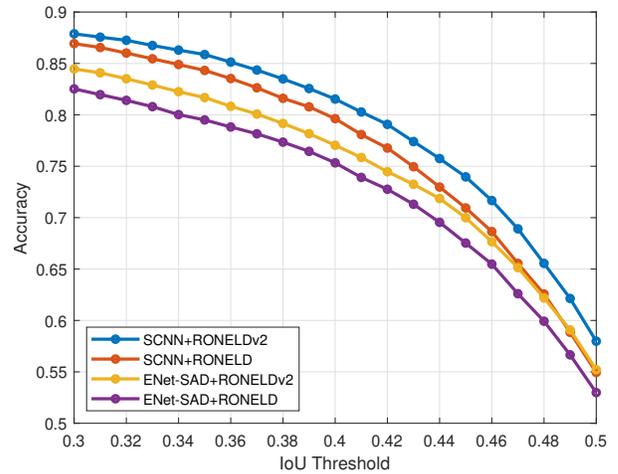}\label{tusimple_graph}
    }
    \vspace{2mm}
    \caption{Comparative accuracy results for \protect\subref{culane_graph} CULane and \protect\subref{tusimple_graph} TuSimple test sets using CULane-trained state-of-the-art deep learning models, RONELD and RONELDv2, between 0.3 to 0.5 IoU thresholds at 0.01 IoU threshold intervals.}\label{graphFig}
\end{figure}

We use the CULane-trained models to generate probabily map outputs on the lane images from the CULane and TuSimple test sets and use the method outlined in~\cite{Pan2018, Hou2019, Chng2020} to generate lane marking predictions for the SCNN and ENet-SAD models respectively. For the SCNN+RONELD, ENet-SAD+RONELD, SCNN+RONELDv2, and ENet-SAD+RONELDv2 methods, we run the corresponding method on the probability map outputs of the respective models to generate lane marking predictions. We compare the lane marking predictions with the ground truth lane marking provided with the datasets to calculate the accuracy performance of the methods which we list in Tables \ref{CULaneTable} and \ref{TuSimpleTable} and display using graphs in Fig.~\ref{graphFig}.

\begin{table*}
    \setlength{\tabcolsep}{1em}
	\small
	\caption{Comparative accuracy results and recorded runtimes using different lane tracking methods presented together with the SCNN and ENet-SAD lane detection models in~\cite{Pan2018,Hou2019} respectively, separately in RONELD~\cite{Chng2020}, and this paper on the CULane test set. The highest accuracy measure on each model at each IoU threshold and lowest runtime is presented in bold.} \label{CULaneTable}
	\centering
	\begin{tabular}{| c | c c | c | c c | c |}
		\hline
        Model:
        & \multicolumn{3}{c|}{SCNN}
        & \multicolumn{3}{c|}{ENet-SAD}\\
		\hline
		IoU Threshold
		& Original
		& RONELD
		& RONELDv2 (Ours)
		& Original
		& RONELD
		& RONELDv2 (Ours)\\
		\hline
		\hline
		0.3
		& 0.812
		& 0.826
		& \textbf{0.832}
		& 0.823
		& 0.832
		& \textbf{0.835}\\
		0.4
		& 0.762
		& 0.789
		& \textbf{0.797}
		& 0.778
		& 0.799
		& \textbf{0.804}\\
		0.5
		& 0.629
		& 0.703
		& \textbf{0.714}
		& 0.655
		& 0.729
		& \textbf{0.737}\\
		\hline
		\hline
		Runtime (ms) 
		& -
		& 2.82 (5.68*)
		& \textbf{1.48}
		& -
		& 3.59 (6.29*)
		& \textbf{2.16}\\
		\hline
	\end{tabular}\\
	\vspace{1.5mm}
	\raggedright
	\begin{footnotesize}{* : original RONELD implementation from~\cite{Chng2020} using a different CPU (Intel Core i9-9900K)}
    \end{footnotesize}
\end{table*}

\begin{table*}[ht]
    \setlength{\tabcolsep}{1em}
    \renewcommand{\arraystretch}{1.05}
	\small
	\caption{Comparative accuracy results and recorded runtimes using different lane tracking methods presented together with the SCNN and ENet-SAD lane detection models in~\cite{Pan2018,Hou2019} respectively, separately in RONELD~\cite{Chng2020}, and this paper on the TuSimple test set, which is a cross-dataset validation using CULane-trained models. The highest accuracy measure on each model at each IoU threshold and lowest runtime is presented in bold.} \label{TuSimpleTable}
	\centering
	\begin{tabular}{|c | c c | c | c c | c |}
		\hline
        Model:
        & \multicolumn{3}{c|}{SCNN}
        & \multicolumn{3}{c|}{ENet-SAD}\\
		\hline
		IoU Threshold
		& Original
		& RONELD
		& RONELDv2 (Ours)
		& Original
		& RONELD
		& RONELDv2 (Ours)\\
		\hline
		\hline
		0.3
		& 0.625
		& 0.869
		& \textbf{0.879} 
		& 0.608
		& 0.825
		& \textbf{0.845} \\
		0.4
		& 0.470
		& 0.796
		& \textbf{0.815}
		& 0.502
		& 0.753
		& \textbf{0.770} \\
		0.5
		& 0.238
		& 0.549
		& \textbf{0.580}
		& 0.341
		& 0.530
		& \textbf{0.552} \\
		\hline
		\hline
		Runtime (ms) 
		& -
		& 2.07 (2.80*)
		& \textbf{1.00} 
		& -
		& 2.76 (3.55*)
		& \textbf{1.57}\\
		\hline
	\end{tabular}\\
	\vspace{1.5mm}
	\raggedright
	\begin{footnotesize}{* : original RONELD implementation from~\cite{Chng2020} using a different CPU (Intel Core i9-9900K)}
    \end{footnotesize}
\end{table*}

%graph of results

\subsection{Results}

\subsubsection{Accuracy performance} We generally see an increase in the accuracy performance using our proposed method beyond those achieved by RONELD. In particular, we see a greater increase in accuracy performance on the TuSimple cross-dataset validation test and at higher IoU thresholds as well, increasing from 0.549 to 0.580 at the highest 0.5 IoU threshold using SCNN on the TuSimple test set. This greater improvement in performance at higher IoU thresholds and on the TuSimple test set can generally be explained by the ability of RONELDv2 to refine the lane parameters using its improved use of preceding frame information. RONELDv2 was able to detect these more accurate lane parameters through the use of its lane merging step and lane point variance information to improve on the lanes detected by the original RONELD method, however these two steps in particular are unable to make changes to lanes which are far apart and hence not matched together. As a result, these steps help RONELDv2 detect lanes which better meet the higher IoU thresholds but have a less significant impact on poorly detected lanes that are far from the ground truths and hence do not meet even the lowest 0.3 IoU threshold. RONELDv2 also had a more significant improvement in accuracy performance on the TuSimple dataset due to the greater reliance on preceding frame information to obtain accurate lane information on this dataset, where RONELDv2 is able to make better use of this preceding frame information. This can be seen by the relatively low initial performance of the deep learning lane detection models on the TuSimple cross-dataset validation test and in our ablation studies on the effectiveness of RONELDv2 methods.

%Noticebly, In the TuSimple test set, although the performance on the 0.3 IoU threshold surpasses that of the CULane test set there is still a more significant room for improvement as the TuSimple is a relatively simple dataset which results in a potentially lower Bayes error rate and hence a higher possible accuracy. At the higher IoU thresholds, there is lower accuracy performance which provides the greater room for improvement. This greater room for improvement is exploited by our method that results in a more significant increase in accuracy performance on the TuSimple test set and on higher IoU thresholds.  

\subsubsection{Runtime} To test the runtime for our proposed method, we recorded the mean runtime of the method on all images on the dataset using a single Ryzen 7 4800H CPU. We record runtime of our method in Tables~\ref{CULaneTable} and ~\ref{TuSimpleTable} together with the runtimes from~\cite{Chng2020} for comparison. In our experiments, we implemented some additional optimizations on the RONELD method which led to a lower runtime using the RONELD method despite the slower CPU (mean runtime for the SCNN CULane dataset was 5.68ms for Intel Core i9-9900K~\cite{Chng2020} and 7.30ms for Ryzen 7 4800H using the same original implementation of RONELD), which we recorded for RONELD as well. These additional optimizations include better use of just-in-time compilation and a closed form expression for the integral in the calculation of the lane distance instead of using numerical integration methods for calculating the distance between the predicted straight lanes. From our experiments, when comparing between RONELD and RONELDv2 methods that use the new optimization, it is observed that there is an up to two fold decrease in the runtime for RONELDv2 when compared with RONELD. This can be attributed to the improved weight system used in our new method which reduces the maximum weight of stored lanes. This lower maximum weight allows these lanes to be cleared more rapidly from the stored cache of previous lanes that current lanes are compared to when searching for the matching previous lane, thereby reducing the amount of processing needed for the lane merging and tracking steps of our method. The greater absolute decrease in the runtime on the CULane dataset \textit{vis a vis} the TuSimple dataset can be attributed to the greater number of lanes detected on the CULane dataset as the models used were trained on the CULane train set. Comparing the runtimes on the SCNN and ENet-SAD models, the difference in runtime can be attributed to the different sizes of the probability map outputs from the model, with the ENet-SAD model used outputting a $976 \times 351$ probability map while the SCNN model used outputs a $800 \times 288$ probability map. Rescaling the ENet-SAD model output probability map to size $800 \times 288$ results in a runtime of 1.62ms on the CULane dataset which is similar to the runtime of 1.48ms for the SCNN model on the same dataset.

\subsection{Ablation study}

%alpha (preceding frame tracking) table
\begin{table}
  \begin{center}
      \caption{Comparison of accuracy results of SCNN + RONELDv2 with different $\alpha$ values at 0.3 and 0.5 IoU thresholds on the CULane and TuSimple test sets.}\label{PFTtable}
      \vspace{-2mm}
      \textbf{CULane}
      \vspace{1mm}
      
      \begin{tabular}{c|c c c c}
        \hline
        IoU Threshold & $\alpha = 0.25$ & $\alpha = 0.5$ & $\alpha = 0.75$ & $\alpha = 1$\\
        \hline
        0.3 & 0.783 & \textbf{0.832} & 0.831 & 0.825 \\
        0.5 & 0.644 & \textbf{0.714} & 0.713 & 0.712 \\
        \hline
      \end{tabular}
      
      \vspace{3mm}
      \textbf{TuSimple}
      \vspace{1mm}
      
      \begin{tabular}{c|c c c c}
        \hline
        IoU Threshold & $\alpha = 0.25$ & $\alpha = 0.5$ & $\alpha = 0.75$ & $\alpha = 1$\\
        \hline
        0.3 & \textbf{0.887} & 0.879 & 0.869 & 0.809 \\
        0.5 & \textbf{0.582} & 0.580 & 0.571 & 0.544 \\
        \hline
      \end{tabular}
  \end{center}
\end{table}

\subsubsection{$\alpha$ value} To measure the impact of the exponentially weighted moving average method for calculating lane weights and to determine the impact of different $\alpha$ values for this step, we test different $\alpha$ values from 0.25 to 1 (inclusive) at 0.25 intervals, where an alpha value of 1 corresponds to no weight placed on the previous $\Omega_{f-1}$ value, though we still retain the lane merge step to merge lane parameters covered in section \ref{lanemergesection} if the lanes are matched, with the results recorded in Table \ref{PFTtable}. It is observed that performance on the TuSimple test set increases as $\alpha$ value decreases, which reflects the positive impact that preceding frame tracking has in this cross-dataset validation test by using information from preceding frames to supplement the relatively weak performance in the current frame. On the other hand, on the CULane test set, performance peaks at $\alpha = 0.5$ which corresponds to an equal weight placed on the weight of the lane in the current frame and on the exponentially weighted average value from the previous frame. $\alpha = 0.5$ value was selected due to the maximum performance achieved on the CULane test set and diminishing returns achieved on the TuSimple test set by decreasing $\alpha$ further.

\subsubsection{Lane match distance} To determine the optimal threshold for the length (measured in standard deviation of the lane points) between lanes that are matched together in our lane tracking process, we test different lane distance values based on the standard deviation of the lane point locations calculated in section \ref{variancesection} and list the results in Table \ref{lanematchtable}. We tested maximum lane match distances set at 0, 1$\sigma$, 2$\sigma$, and 3$\sigma$, with a lane match distance of 0 corresponding to lanes being matched only when their lane parameters match exactly which is rare in practice. For this ablation study, the lanes that are not matched are still stored for subsequent frames, with their weights undergoing exponential decay in line with the exponentially weighted moving average method used to calculate lane weights, and this is intended to isolate the impact of the lane matching step. From our experiments, it is observed that a lane match distance of $2\sigma$ provides an optimal combined result across the CULane and TuSimple test sets, with minor decrease in results of less than 0.01 accuracy compared to using a $1\sigma$ or $3\sigma$ lane match distance at the 0.3 IoU threshold but a more significant increase in accuracy compared to the other lane match distances at the 0.5 IoU threshold
.

\subsubsection{Lane merge} To verify the usefulness of merging the lane parameters in our method, we test the difference in accuracy performance with and without the lane merge step and list the results in Table~\ref{lanemergetable}. It is observed that the lane merge step generally increases the accuracy performance of the method. In particular, it has a greater impact at the higher 0.5 IoU threshold which is as expected as the lane merge step helps to finetune the lane parameters of lanes that are relatively close to each other due to the prior lane match step. As such, the merging of the lane parameters generally makes small adjustments to lanes that are already within a close vicinity to each other, which helps them reach a higher IoU threshold but is unable to improve the result of a poorly detected lane that is far from the ground truth lane marking as those lanes would not have been merged together and have no impact on each other.

\begin{table}
  \begin{center}
      \caption{Comparison of accuracy results of SCNN + RONELDv2 with different lane match distance values at 0.3 and 0.5 IoU thresholds on the CULane and TuSimple test sets.}\label{lanematchtable}
      \vspace{-2mm}
      \textbf{CULane}
      \vspace{1mm}
      
      \begin{tabular}{c|c c c c}
        \hline
        IoU Threshold & $0$ & $1\sigma$ & $2\sigma$ & $3\sigma$\\
        \hline
        0.3 & 0.832 & \textbf{0.832} & 0.832 & 0.832 \\
        0.5 & 0.712 & 0.713 & \textbf{0.714} & 0.712 \\
        \hline
      \end{tabular}
      
      \vspace{3mm}
      \textbf{TuSimple}
      \vspace{1mm}
      
      \begin{tabular}{c|c c c c}
        \hline
        IoU Threshold & $0$ & $1\sigma$ & $2\sigma$ & $3\sigma$\\
        \hline
        0.3 & 0.872 & 0.876 & 0.879 & \textbf{0.880} \\
        0.5 & 0.567 & 0.573 & \textbf{0.580} & 0.578 \\
        \hline
      \end{tabular}
  \end{center}
\end{table}

%lane merge table
\begin{table}
  \begin{center}
      \caption{Accuracy results of SCNN + RONELDv2 with and without lane merge (LM) step on the CULane and TuSimple test sets at 0.3 and 0.5 IoU thresholds.}\label{lanemergetable}
      
      \begin{tabular}{c|c c c c}
        \hline
        \multirow{2}{*}{IoU Threshold} &
          \multicolumn{2}{c}{CULane} &
          \multicolumn{2}{c}{TuSimple} \\
          \cline{2-5}
          & {w/o LM} & {w/ LM} & {w/o LM} & {w/ LM}\\
          \hline
        0.3 & 0.831 & \textbf{0.832} & 0.875 & \textbf{0.879} \\
        0.5 & 0.711 & \textbf{0.714} & 0.569 & \textbf{0.580} \\
        \hline
    \end{tabular}
  \end{center}
\end{table}

\section{Conclusion}
In this paper, we have presented an improved lane tracking method that improves on a previous method by finding the lane point variance, using an exponentially weighted moving average method to weigh lanes, and merging lane parameters. We have demonstrated the usefulness of our method on the CULane and TuSimple test sets using the SCNN and ENet-SAD lane detection models with an increase in accuracy performance and an up to two-fold decrease in the runtime of our method. This shows the usefulness of lane tracking methods in improving the accuracy performance of deep learning lane detection methods, particularly on cross-dataset validation tests.

%\appendices
%\section{Proof of the ...}
%Appendix one text goes here.

%\ifCLASSOPTIONcaptionsoff
%  \newpage
%\fi

\bibliographystyle{IEEEtran}
\bibliography{main}

\end{document}